%% file: main.tex
\documentclass[10pt,twocolumn,letterpaper]{article}

\usepackage[pagenumbers]{cvpr} 

\usepackage{graphicx}
\usepackage{amsmath}
\usepackage{amssymb}
\usepackage{booktabs}
\usepackage{xcolor}
\usepackage{algorithm}
\usepackage{algorithmic}
\usepackage{colortbl}
\usepackage{footnote}

\usepackage[pagebackref,breaklinks,colorlinks]{hyperref}

\usepackage[capitalize]{cleveref}
\crefname{section}{Sec.}{Secs.}
\Crefname{section}{Section}{Sections}
\Crefname{table}{Table}{Tables}
\crefname{table}{Tab.}{Tabs.}

\newcommand{\vh}{\mathcal{V}}

\newcommand{\speedup}{2-8x}
\newcommand{\numlines}{30}

\def\cvprPaperID{7886} 
\def\confName{CVPR}
\def\confYear{2022}

\begin{document}
\input{macros.tex}

\title{VaxNeRF: Revisiting the Classic for Voxel-Accelerated Neural Radiance Field}

\author{Naruya Kondo\\
University of Tsukuba\\
{\tt\small n-kondo@digitalnature.slis.tsukuba.ac.jp}
\and
Yuya Ikeda\\
University of Tokyo\\
{\tt\small y-ikeda0124@g.ecc.u-tokyo.ac.jp}
\and
Andrea Tagliasacchi\\
Google Brain, University of Toronto\\
{\tt\small atagliasacchi@google.com}
\and
Yutaka Matsuo\\
University of Tokyo\\
{\tt\small matsuo@weblab.t.u-tokyo.ac.jp}
\and
Yoichi Ochiai\\
University of Tsukuba\\
{\tt\small wizard@slis.tsukuba.ac.jp}
\and
Shixiang Shane Gu\\
Google Brain\\
{\tt\small shanegu@google.com}
}
\maketitle


\begin{abstract}
Neural Radiance Field (NeRF) is a popular method in data-driven 3D reconstruction. Given its simplicity and high quality rendering, many NeRF applications are being developed. However, NeRF's big limitation is its slow speed. Many attempts are made to speeding up NeRF training and inference, including intricate code-level optimization and caching, use of sophisticated data structures, and amortization through multi-task and meta learning. In this work, we revisit the basic building blocks of NeRF through the lens of classic techniques before NeRF. We propose Voxel-Accelearated NeRF (VaxNeRF), integrating NeRF with visual hull, a classic 3D reconstruction technique only requiring binary foreground-background pixel labels per image. Visual hull, which can be optimized in about 10 seconds, can provide coarse in-out field separation to omit substantial amounts of network evaluations in NeRF. We provide a clean fully-pythonic, JAX-based implementation on the popular JaxNeRF codebase, consisting of only about 30 lines of code changes and a modular visual hull subroutine, and achieve about \speedup~faster learning on top of the highly-performative JaxNeRF baseline with zero degradation in rendering quality. With sufficient compute, this effectively brings down full NeRF training from hours to 30 minutes. We hope VaxNeRF -- a careful combination of a classic technique with a deep method (that arguably replaced it) -- can empower and accelerate new NeRF extensions and applications, with its simplicity, portability, and reliable performance gains. Codes are available at \url{https://github.com/naruya/VaxNeRF}.
\end{abstract}

\input{sections/1_intro}

\input{sections/2_relworks}

\input{sections/3_prelim}

\input{sections/4_method}
\input{sections/5_results}
\input{sections/6_conclusion}


{\small
\bibliographystyle{ieee_fullname}
\bibliography{main}

}
\newpage
\appendix
\input{sections/9_appendix}

\end{document}

%% file: macros.tex
\newcommand\estimate[1]{\hat{#1}}
\newcommand{\ray}{\mathbf{r}}
\newcommand{\raybatch}{\mathcal{R}}
\newcommand{\position}{\mathbf{x}}
\newcommand{\direction}{\mathbf{d}}
\newcommand{\col}{c}
\newcommand{\Col}{C}
\newcommand{\truecol}{C}

\definecolor{colorthird}{rgb}{0.83, 0.93, 1.00} 
\definecolor{colorsecond}{rgb}{0.87, 0.97, 0.83} 
\definecolor{colorfirst}{rgb}{1.00, 0.97, 0.83} 
\definecolor{coloryes}{rgb}{0.20, 0.80, 0.00} 

\newcommand{\textfirstb}{\colorbox{colorfirst}}
\newcommand{\textsecondb}{\colorbox{colorsecond}}
\newcommand{\textthirdb}{\colorbox{colorthird}}

\newcommand{\cellfirst}{\cellcolor{colorfirst}}
\newcommand{\cellsecond}{\cellcolor{colorsecond}}
\newcommand{\cellthird}{\cellcolor{colorthird}}

\newcommand{\Yes}{{\color{coloryes} \textbf{Yes}}}
\newcommand{\No}{{\color{red} \textbf{No}}}
\newcommand{\Lower}{{\color{red} Lower}}
\newcommand{\Required}{{\color{red} Required}}

%% file: sections/1_intro.tex
\section{Introduction}
\label{sec:intro}

\input{figures/top}



Neural radiance field (NeRF)~\cite{mildenhall2020nerf} has revolutionized high-resolution 3D reconstruction from images. 
Many extensions and applications quickly ensued, including dynamic scene rendering~\cite{Li20arxiv_nsff, Gafni_2021_CVPR, tretschk2021nonrigid, 2021narf, peng2021animatable, park2021hypernerf}, controllable relighting~\cite{boss2021nerd, nerv2021, Wizadwongsa2021NeX, nerfactor}, latent appearance and shape priors~\cite{Schwarz2020NEURIPS, chanmonteiro2020pi-GAN, trevithick2020grf}, scene composition~\cite{Niemeyer2020GIRAFFE, guo2020osf, Ost_2021_CVPR, yang2021objectnerf}, and pose estimation~\cite{yen2020inerf, Sucar:etal:ICCV2021, meng2021gnerf, lin2021barf, jeong2021self}. 

These works have potential to make substantial impacts not only in computer vision and computer graphics, but also in robotics~\cite{li20213d,adamkiewicz2021vision}, virtual reality (VR) systems~\cite{bruno20103d, hirose2005virtual, walczak2006virtual, nayyar2018virtual}, and medical imaging~\cite{udupa19993d}.

Despite NeRF's high quality 3D renderings, a key bottleneck has been slow training and inference speeds where the original implementation took days to render an object and tens of seconds to render a view angle~\cite{mildenhall2020nerf}. Given this, follow up work looked to accelerate NeRF, including caching with special data structures~\cite{hedman2021snerg, garbin2021fastnerf, yu2021plenoctrees}, amortization through meta learning and multi-task learning~\cite{tancik2020meta, bergman2021metanlr}, or re-implementation optimized for parallel computing~\cite{reiser2021kilonerf, jaxnerf2020github}, where most approaches focus on inference time speedups~\cite{reiser2021kilonerf, hedman2021snerg, garbin2021fastnerf, yu2021plenoctrees, autoint}. Notably, many of these approaches take NeRF as given and add extra complexities to enhance its performances, rather than fundamentally changing its design details, except a few~\cite{kaizhang2020, autoint, liu2020neural}.

Importantly, prior to NeRF there were many kinds of classic 3D reconstruction techniques~\cite{laurentini1994visual, schoenberger2016sfm, schoenberger2016mvs, westoby2012structure}. Instead of taking NeRF for granted and training end-to-end models,  we revisit the basic building blocks of NeRF and ask: can classic techniques be combined with NeRF in meaningful ways? In this work, we focus on the points sampling in ray-color integration, which is a fundamental bottleneck to NeRF's speed and is generally addressed through hierarchical sampling with coarse and fine models~\cite{mildenhall2020nerf}.  Visual hull~\cite{laurentini1994visual}, a classic 3D reconstruction technique, can obtain an approximate bounding volume, with only a binary foreground-background segmentation mask per image, which is easily obtainable  through thresholding or recent off-the-shelf segmentation models~\cite{BGMv2, Germer2020, sun2021sim}. Visual hull can be fitted for 100 images in about 10 seconds, using our JAX-based implementation, and using this coarse voxel model, we can eliminate most of the expensive point evaluations that are not inside the object. We term our method Voxel-Accelerated NeRF (VaxNeRF), which can eliminate the need for an extra coarse model, and can provide a consistent \speedup~faster training speedup on top of the highly-performative JaxNeRF~\cite{jaxnerf2020github} with a portable JAX-based visual hull file and only around \numlines~lines of modification to the main code.

We perform extensive evaluations using NeRF-Synthetic datasets, along with more difficult datasets of NSVF~\cite{liu2020neural}, to ensure that our observed speedups are consistent across various scenes without degradation in the final rendering quality (measured with PSNR and SSIM). We emphasize those scenes with non-convexities or transparencies, where visual hull alone certainly cannot provide ground-truth in-out volumes, but still observe that VaxNeRF provides similar degrees of speedups. On average VaxNeRF speeds up (Jax)NeRF training by \speedup, with the same and sometimes higher final rendering accuracy. To our surprise, VaxNeRF, despite being a much simpler method to implement on top of NeRF or NeRF-based algorithms, also outperforms recent NSVF~\cite{liu2020neural}, as well as PlenOctrees~\cite{yu2021plenoctrees} with respect to PSNRs, while providing better or comparable training speeds.

We hope that VaxNeRF, with its simplicity, portability, and reliable performance gains, can accelerate and empower many future NeRF extensions and applications. Extrapolating on the 256-TPU performance reported on JaxNeRF~\cite{jaxnerf2020github}, VaxNeRF brings down full NeRF training from 2.5 hours to 30 minutes, only requiring additional binary foreground-background pixel labels. Our method, which simply combines a classic technique with a recent end-to-end technique, demonstrates the importance of revisiting core building blocks of popular deep learning models as well as classical techniques prior to them, as sometimes, the best of both worlds is a careful combination of the two.    


%% file: figures/top.tex
\begin{figure}[t]
  \centering

  \begin{subfigure}{0.49\linewidth}
    \includegraphics[width=1.\linewidth]{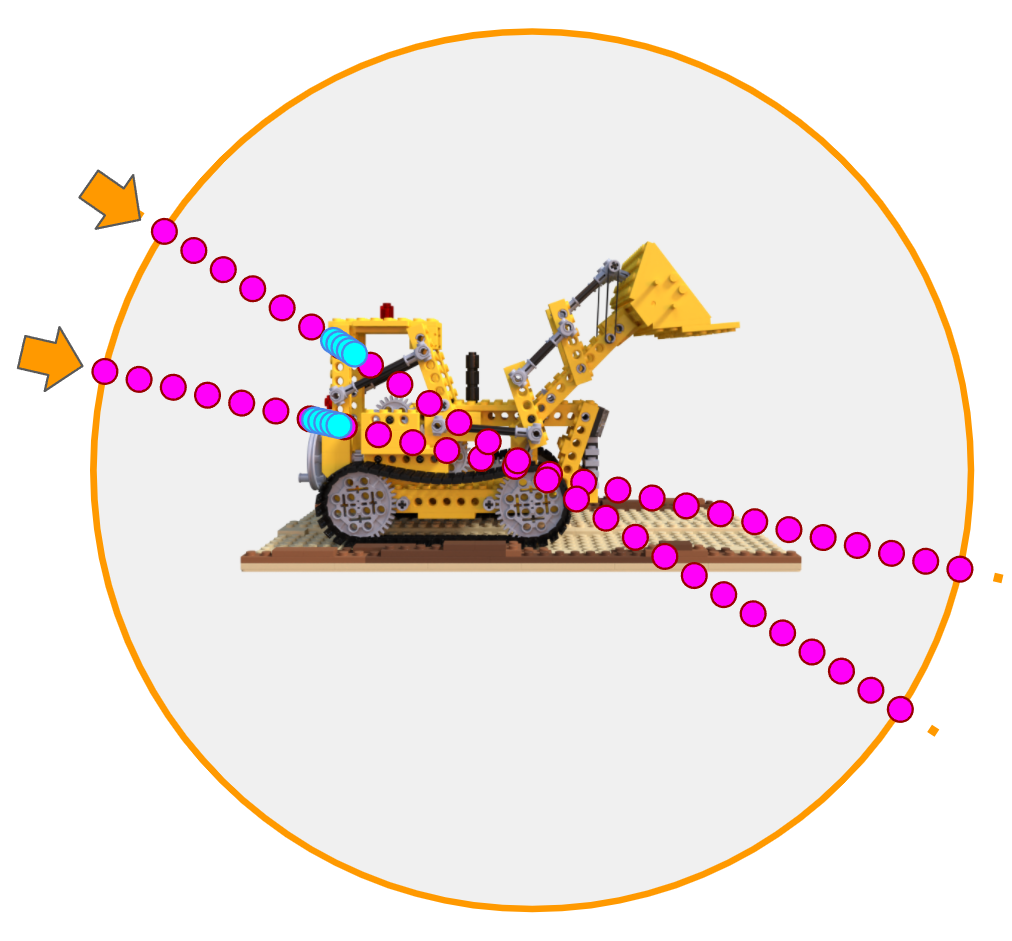}
    \caption{NeRF}
    \label{fig:top-a}
  \end{subfigure}
  \hfill
  \begin{subfigure}{0.49\linewidth}
    \includegraphics[width=1.\linewidth]{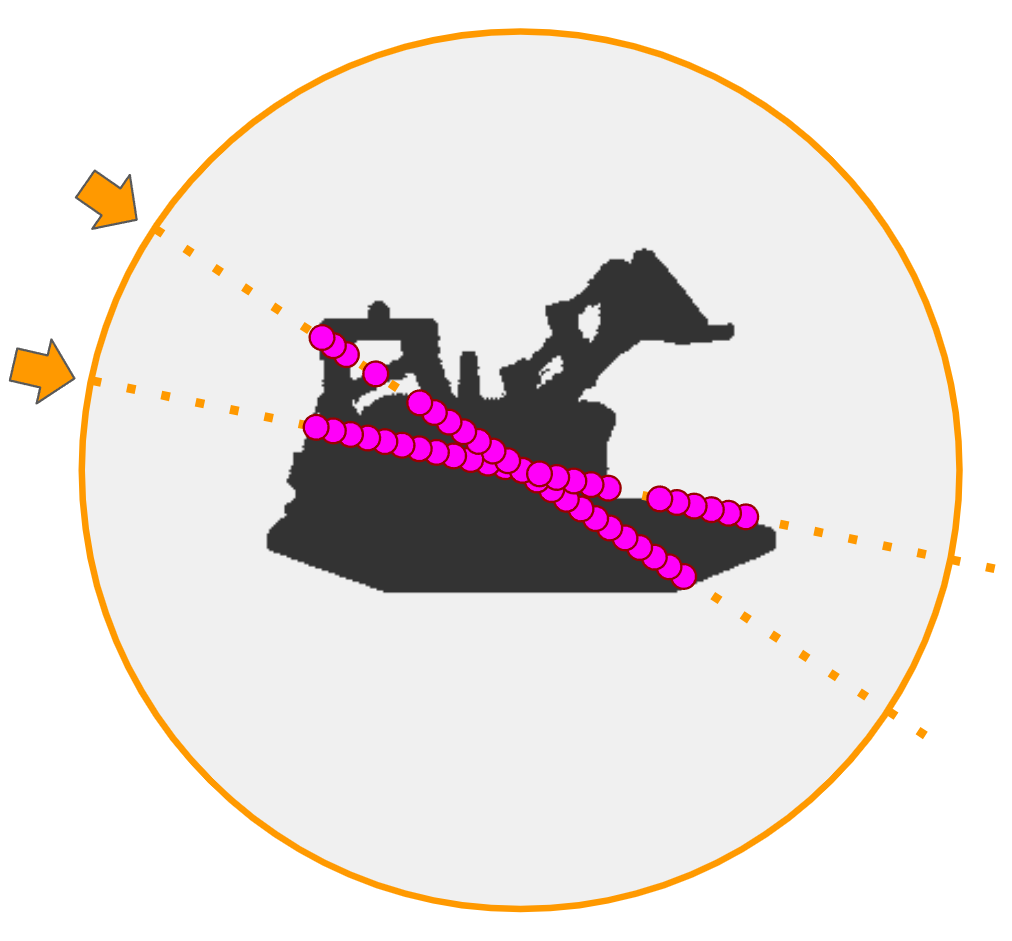}
    \caption{VaxNeRF}
    \label{fig:top-b}
  \end{subfigure}

  \caption{NeRF (a) uses hierarchical sampling, i.e., a mixture of coarse (pink) and fine (light blue) sampling, to estimate the color from a specified viewpoint. Instead, VaxNeRF (b) samples only the inside of the bounding volume uniformly and densely. The orange circle shows the area bounded by the minimum and maximum distances of the rays, and the black shade \mbox{in (b)} shows the automatically obtained bounding volume.}
  \label{fig:topimage}
\end{figure}

%% file: sections/2_relworks.tex
\section{Related Works}
\label{sec:relwork}


\subsection{Novel View Synthesis}

The task of synthesizing novel views of a scene given multiple images has been well studied for a long time, with various methods that divide into two categories: 3D reconstruction methods and viewpoint interpolation methods.

\paragraph{3D Reconstruction} The voxel-based methods~\cite{laurentini1994visual, kutulakos2000theory, seitz1999photorealistic} are very primitive methods for 3D reconstruction, which infer the occupancy or color of each point in grid space using the silhouette of the image or the viewpoint invariances of the color. Along with voxel representation, meshes are another commonly used representation\cite{waechter2014, buehler01, debevec96, wood00} since they are easy to use and can store information efficiently, but it is difficult to capture or optimize detailed geometry and topologies. MVS~\cite{kar2017learning, tulsiani2017multi} is an example of practical applications of 3D reconstruction form images, and is commonly used in 3D modeling work today. Recently, several deep learning-based approaches are proposed to continuously represent the 3D space ~\cite{Lombardi:2019, trevithick2020grf, Lior2021volume, mildenhall2020nerf}. Neural Volumes~\cite{Lombardi:2019} performs 3D reconstruction by 3D deconvolution from accumulated image features, and then synthesize image from novel views by ray marching. Similarly, NeRF\cite{mildenhall2020nerf} uses ray marching, but instead of using 3D deconvolution, it learns a model that infers color and density given the coordinates and viewing direction along the camera ray. NSVF~\cite{liu2020neural} is akin to a combination of NeRF and Neural Volumes, and speeds up the inference of NeRF by allowing features to be stored in a voxel grid. And lastly, there are other methods of 3D reconstruction using implicit functions such as the signed distance function ~\cite{Lior2021volume, trevithick2020grf}.

\paragraph{Viewpoint Interpolation} Multi plane images~\cite{zhou2018stereo, single_view_mpi} and light field interpolation~\cite{davis2012unstructured, levoy1996light, gortler1996lumigraph, mildenhall2019llff} are image-based approaches for novel view synthesis without 3D reconstruction. When the postures of novel viewpoints can be limited, such as forward-facing scenes, these representations are very light-weight, accurate and effective, but otherwise, their use case is limited. Additionally, there are novel view synthesis methods use convolution-based deep generative models\cite{eslami2018neural, HoloGAN2019}. While these are good at generating photo-realistic images at a reasonable speed, they have a problem that the geometry are prone to collapsing. 


\subsection{NeRF Accelerations}

NeRF is attractive because of its ability to learn and represent precise shapes and complex reflections. However, NeRF's training and inference (rendering) speed are very slow~\cite{mildenhall2020nerf}\footnote{DVGO~\cite{cheng2021direct} is a concurrent paper to ours, which proposes a method of storing features in voxels (similar to NSVF) and achieves 49-183x faster learning than NeRF. However, their implementation is completely different from NeRF and therefore difficult to combine with existing NeRF research. Additionally, DVGO requires a larger memory footprint, where the model size totals 650M, compared to 6M in NeRF and 12M in our method (6M for model, 6M for binary voxels).}. Many researchers have proposed methodologies to overcome this problem. ~\cite{yu2021plenoctrees, garbin2021fastnerf, hedman2021snerg, reiser2021kilonerf, rebain2021derf} are great methods for speeding up the NeRF inference. While the original NeRF took about 30 seconds to render one image, these methods achieve real-time rendering by pre-computing and storing colors and densities in scene space, or by dividing the NeRF model into smaller models for better parallel inference. Another way to speeding up NeRF on a new scene is to incorporate priors learned from a dataset of similar scenes, which can be accomplished by conditioning on predicted images features ~\cite{trevithick2020grf,yu2020pixelnerf,wang2021ibrnet,chen2021mvsnerf} or meta-learning ~\cite{tancik2020meta}. While these methods are effective for simple scenes or rough 3D reconstructions, these are not yet able to be used to generate complex scenes in the NeRF paper to the same quality. TermiNeRF~\cite{piala2021terminerf} is a more direct way to speed up NeRF, where along with the usual NeRF model, it simultaneously trains a sampling network that predicts the locations of points to be sampled on a ray. Although it is claimed to speed up fine-tuning to a scene with the same shape but different look and lighting that has been learned once, it does not mention speeding up training itself. AutoInt\cite{autoint} modifies the architecture of the NeRF to be able to calculate color and density by definite integration, however, the learning speedup is limited to 1.8x. Another approach for speeding up is to focus only on the object surfaces for 3D reconstruction and rendering~\cite{Oechsle2021ICCV, wang2021neus, yifan2020isopoints, lior2021volsdf, Oechsle2021ICCV}, but it is extremely difficult to correctly estimate the shapes for objects with transparent parts or fuzzy objects whose surface is difficult to estimate. Our method, on the other hand, achieves faster training without narrowing down applicable domains or sacrificing rendering quality.

\input{tables/relworks}

\subsection{NeRF and 3D Geometry Integration}

There are several work that expand NeRF using geometry data inferred from images or given as a supervised information. ~\cite{Wang2021MirrorNeRFON, niemeyer2020differentiable} use a rough bounding volume obtained from visual hull~\cite{laurentini1994visual} to measure occupancy loss to get a clear estimate of the object surface, but they do not focus on obtaining significant speed-ups. NeLF~\cite{sun2021nelf} is an intricate, highly-customized method for portrait view synthesis that combines NeRF and NeuralVolumes. NeLF uses visual hull to reduce the number of sampling points on a ray, but does not study how effective it is alone or in the most basic form of NeRF. NeuRays~\cite{liu2021neuray} is an image-based rendering method that utilizes the surface information of objects obtained by Colmap~\cite{schoenberger2016mvs, schoenberger2016sfm}. DoNeRF~\cite{neff2021donerf} and DS-NeRF~\cite{deng2021depth} use depth images to reduce the number of required observation images and speed up training, but our method achieves faster training without such depth oracle.

%% file: tables/relworks.tex
\newcommand{\kome}{\textreferencemark}

\begin{savenotes}
\begin{table*}[!t]

\centering
\begin{tabular}{lccccccc}
\toprule
                          & {\small fast rendering} & {\small fast training} & {\small quality} & {\small transparency} & depth & {\small representation} & \begin{tabular}{c}
{\small code} \\ {\small modification}
\end{tabular} \\
\midrule

NeRF~\cite{mildenhall2020nerf}            & ($\approx$ 0.05fps)  & ---            & Good         & Yes & No & NR & ---     \\ 
KiloNeRF~\cite{reiser2021kilonerf}        & $\approx$ 1000x      & N/A             & $\sim$ Good  & Yes & No & NR & 1000s+ \\
SNeRG~\cite{hedman2021snerg}              & $\approx$ 1600x      & N/A             & $\sim$ Good  & Yes & No & NR + Cache & 1000s+\\
FastNeRF~\cite{garbin2021fastnerf}        & $\approx$ 4000x      & N/A             & $\sim$ Good  & Yes & No & Cache & 1000s+\\
PlenOctrees~\cite{yu2021plenoctrees}      & $\approx$ 3000x      & $\approx$ 4x   & $\sim$ Good  & Yes & No & SH & $\approx$ 1000s \\
NSVF~\cite{liu2020neural}                 & $\approx$ 10x        & N/A             & Good & Yes & No & NR & 1000s+ \\
AutoInt~\cite{autoint}                    & $\approx$ 10x        & $\approx$ 1.8x & \Lower & Yes & No & NR & 1000s+ \\
TermiNeRF~\cite{piala2021terminerf}       & $\approx$ 14x        & N/A\footnote{TermiNeRF claims to be able to do fast finetuning for different looking scenes with the same geometry, but it does not mention how much faster normal training will be.} & Good            & Yes & No & NR & unknown \\
MetaNLR~\cite{bergman2021metanlr}         & $\approx$ 20x        & $\approx$ 8x & $\sim$ Good & \No  & No & NR &  unknown  \\
DoNeRF~\cite{neff2021donerf}              & 15-78x               & $\approx$ 6x\footnote{Compared to NSVF\cite{liu2020neural}}             & Good & Yes & \Required & NR &  1000s+   \\
DS-NeRF~\cite{deng2021depth}              & N/A                   & 2-6x           & $\sim$ Good  & Yes & \Required & NR & 1000s+     \\
VaxNeRF (Ours)                            & N/A\footnote{We haven't measured properly, but it's a little faster than original NeRF.}      & \speedup  & Good & Yes  & No & NR & $\approx$ 30            \\

\bottomrule
\end{tabular}
\caption{A comparison of efficient NeRF-based methods. \textbf{fast rendering} and \textbf{fast training}: How many times faster the rendering / training are compared to NeRF. ``No" does not mean no speedup at all, but indicates that it is not specifically mentioned in the paper. \textbf{quality}: Rough quality comparison with NeRF as ``Good''. The ``~Good'' indicates that the quality depends on parameters such as compression ratio. \textbf{transparency}: Whether the method can handle objects that cannot be successfully reconstructed by simply estimating the object's surface (e.g. objects with transparency). \textbf{depth}: Whether a depth image is required. \textbf{representation}: The representation of the model that will eventually be created by each method. ``NR'' denotes neural representation, ``cache'' denotes the precomputed values of color and density, and ``SH'' denotes the coefficient of spherical harmonics. \textbf{code modification}: How many lines have been changed/add from the original NeRF implementation. If the code has not been published, displayed as ``unknown''.}
\label{tab:relworks}
\end{table*}


%% file: sections/3_prelim.tex
\section{Preliminaries}
\label{sec:prelim}

Our work builds on the most basic NeRF model with minimal changes. We go over the NeRF basics and introduce visual hull in this section. For more in-depth summaries of NeRFs and extensions, please refer to \cite{dellaert2020neural}.

\subsection{Neural Radiance Fields}
\label{sec:nerf}

Neural radiance fields (NeRF)~\cite{mildenhall2020nerf} are 3D representations that can be rendered from arbitrary viewpoints, capturing continuous geometry and view-dependent appearances. 
The expected color of a camera ray $\mathbf{r}(t) =\mathbf{o} + t\mathbf{d}$ can be estimated by a finite-sample approximation  $\estimate{\Col}(\ray)$:
\begin{align}
    \estimate{\Col}(\ray) &= \sum_{i=0}^{N-1} w_i \, \mathbf{\col}(\ray(t_{i})) \, , \label{eqn:nerfray} \\
    \text{where} \quad 
 w_i &= T_i\big(1 - \exp( -\mathbf{\sigma}(\ray(t_{i})) \delta_{i})\big) \\
  T_i &= \exp\left(-\sum_{j=0}^{i-1}  \mathbf{\sigma}(\ray(t_{j}))   \delta_j\right),
\end{align}
where 
$t_i$ indicates the position of the point to be sampled from the ray, and $\delta_i=t_{t+1}-t_i$ are the distances between point samples.

In NeRF, an MLP parameterizes $\mathbf{\col}(\cdot)$ and $\mathbf{\sigma}(\cdot)$ jointly and is trained to minimize the loss between the ground-truth pixel colors $\truecol(\ray)$ in the images and the predicted colors $\estimate{\Col}$ for a batch of rays $\raybatch$:
\begin{equation}
    \mathcal{L}_{\text{RGB}} = \sum_{\ray \in \mathcal{\raybatch}} \big\lVert{\truecol(\ray) - \estimate{\Col}(\ray)}\big\rVert^2_2
    \label{eqn:nerfloss}
\end{equation}

\paragraph{Hierarchical Sampling}
The NeRF paper~\cite{mildenhall2020nerf} proposes hierarchical sampling to efficiently determine the points $t_i$ to be evaluated for each ray in Eq.~\ref{eqn:nerfray}. Two NeRF models with the same network structure are trained jointly: a coarse model and a fine model.
For the coarse model the points $t_i$ are sampled uniformly, while for the fine model $t_i$ are sampled from an importance-weighted distribution based on their contributions $w_i$ to the final pixel value $\estimate{\Col}$, evaluated using the coarse model. \autoref{fig:top-a} visually shows how the fine model can better allocate samples to important regions, leading to much lower PSNR. 

\subsection{Visual Hull}
\label{sec:vh}

Visual hull~\cite{laurentini1994visual} (or shape-from-silhouette) is a classical method for 3D reconstruction based on a set of images and their camera location information. It assumes that pixels in each captured image can be divided into foreground and background.

The foreground region of an image is created by the projection of a three-dimensional object onto a two-dimensional plane. Therefore, we know that there is absolutely no object on the ray of the camera corresponding to a pixel in the background region, and conversely, there is always an object somewhere on the ray corresponding to a pixel in the foreground region. Therefore, we can narrow down the area where an object may be present using a set of images from multiple angles, and the product of the possible areas where an object is present in each image forms an approximate bounding volume of (a region that completely contains) the object.

Visual hull is a primitive method, and has problems such as inability to correctly restore (carve off) some dented parts of non-convex shapes, but unlike other methods~\cite{kutulakos2000theory, seitz1999photorealistic, chen2021mvsnerf, seitz2006comparison, semerjian2014new, fuhrmann2014mve, galliani2015massively}, it has the advantage of being able to obtain a bounding volume that contains 100\% of the object and therefore has been applied in many ways in classic 3D reconstructions~\cite{szeliski1993rapid, kutulakos1997shape, gortler1996lumigraph}.

While visual hull is a method for estimating shapes and not colors, it can be easily colored by shooting rays from the foreground of the given image to the output voxels of the visual hull.
For visualization, we present a colored output of the visual hull in \autoref{fig: qualitative}, but we did not use this color information in our method.

\end{savenotes}

%% file: sections/4_method.tex
\section{Method}
\label{sec:method}

\begin{figure}[t]
  \centering
  \includegraphics[width=1.\linewidth]{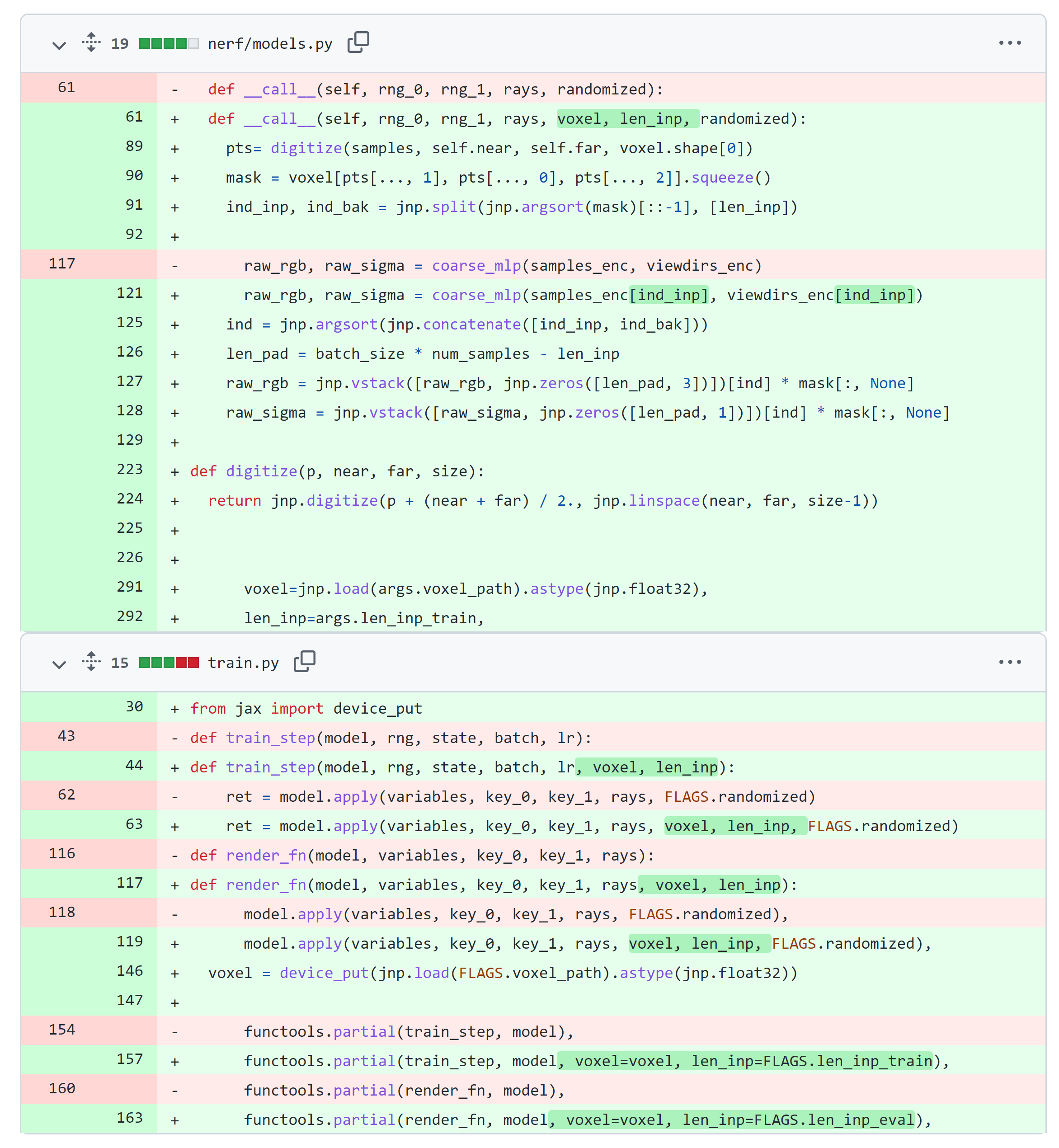}
  \caption{The full code modification to convert from (Jax)NeRF~\cite{mildenhall2020nerf,jaxnerf2020github} to VaxNeRF, visualized with ``git diff". This is not for readability (for algorithmic differences, see Algorithm~\ref{alg:nerf}), but mainly for highlighting how it only takes about \numlines~lines of change to implement VaxNeRF and achieve the speedups, besides a modular visual hull code.}
  \label{fig:commit}
\end{figure}

An obvious yet significant compute inefficiency in NeRF training is its MLP evaluations at the points in space where they have no object and contribute $0$ to $\estimate{\Col}(\ray)$. 
While NeRF addresses this by training a separate coarse model for point selections, we propose a simpler solution:
we modify Eq.~\ref{eqn:nerfray} to only evaluate the points inside of the voxels $\vh$ created by visual hull,
\begin{align}
    \estimate{\Col}(\ray) &= \sum_{i: \ray(t_i)\in \vh} w_i \, \mathbf{\col}(\ray(t_{i})). \label{eqn:vaxnerfray} 
\end{align}
We term it \textbf{Voxel-Accelerated NeRF (VaxNeRF)}. While this is arguably a trivial and intuitive modification to NeRF, VaxNeRF has the following advantages over standard NeRF or other accelerated-NeRF variants:
\begin{itemize}
    \item It is implementable on top of JaxNeRF~\cite{jaxnerf2020github} in about \numlines~lines, fully pythonic and JAX-based (\autoref{fig:commit})
    \item It trains \speedup~faster than JaxNeRF 
    \item It trains without loss of rendering quality measured in PSNR and SSIM
    \item It trains only one network for best performance, as opposed to two in NeRF
\end{itemize}
where the only additional assumption is the access to binary foreground-background masks, which are easily obtainable in most cases as discussed in Section~\ref{sec:setup}. We show how this simple combination of a well-known classical technique and NeRF leads to these impressive empirical performances in Section~\ref{sec:exp_results} (\autoref{fig: qualitative},~\autoref{fig:psnrs},~\autoref{tab:psnrssim},~\autoref{tab:analysis}).


\subsection{Design Choices}
\label{sec:designchoices}

Despite the simplicity and generality of the core VaxNeRF idea, there were a number of key design decisions in VaxNeRF implementation to achieve the best performances.

\paragraph{Choosing visual hull}
As noted briefly in Section~\ref{sec:vh}, visual hull is reliable: it only undercuts spaces (e.g. fail to cut dents in objects), but never overcuts. This property is critical in VaxNeRF, since as long as all object parts are inside $\vh$, it will converge to the same performance as a full NeRF. In Colmap~\cite{schoenberger2016sfm, schoenberger2016mvs} and photogrammetry~\cite{westoby2012structure}, on the other hand, the object surface is estimated using color information that is vulnerable to reflections and other noise, and the estimated surface contains noise and is not suited to be a reliable bounding volume, unlike in visual hull where the estimated volume is 100\% reliable up to the resolution of voxels. We therefore chose visual hull to derive $\vh$ used in VaxNeRF.  

\paragraph{Removing fine model}
As illustrated in~\autoref{fig:topimage}, $\vh$ in VaxNeRF decides point locations based on a binary bounding volume, and the fine model in NeRF decides based on a more direct measure of contribution to the ray color $\estimate{\Col}(\ray)$, i.e. $w_i$, evaluated through the coarse model. Visually it appears that using the fine model of NeRF on top of VaxNeRF could lead to even more speedups; however, through our extensive experiments we found that VaxNeRF performs the best using only a single ``coarse'' model with (dense) uniformly sampling (e.g.~\autoref{fig:psnrs}). This removes the need for fine model and substantially simplifies NeRF training. 




\paragraph{Dilating voxels}
NeRF usually uses 64 and 128 points for coarse and fine sampling respectively~\cite{mildenhall2020nerf}, while our visual hull constructs 400x400x400 voxels. This means that if there is a substructure in the scene that is smaller than 1/64 of the scene size, the ray may ignore that part of the scene even if visual hull was able to reconstruct it. 
In order to avoid this problem, we dilate, i.e. max-pool, the voxel by the amount of cells equal to the scene size divided by the number of point samples, when the sample size is small, e.g. 64.



%% file: sections/5_results.tex
\section{Experiments}
\label{sec:results}

\input{tables/results}
\input{figures/hist}

\begin{figure*}[t]
  \centering
  \includegraphics[width=1.\linewidth]{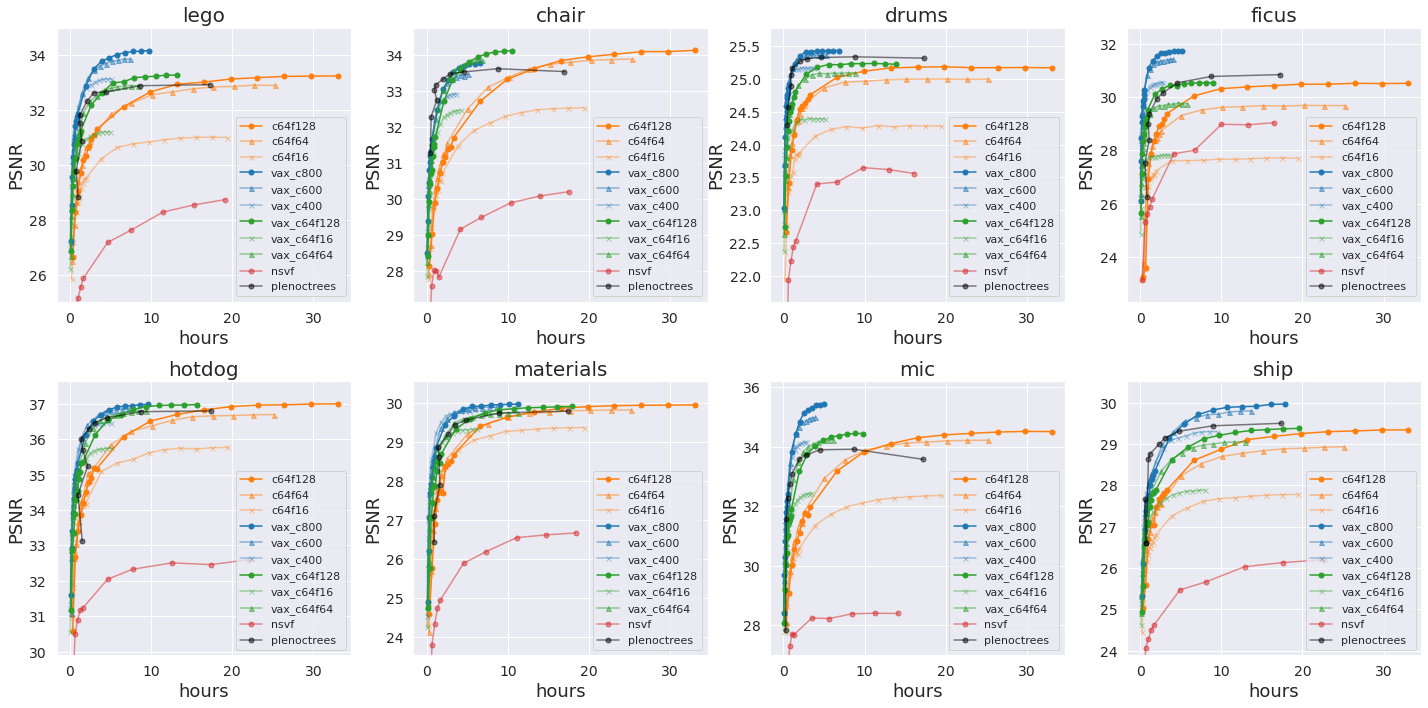}
  \caption{Comparison of training speed. Both VaxNeRF and NeRF trained 1 million iterations. The training of VaxNeRF was \speedup times faster than the original NeRF, and furthermore, vax\_c600 and vax\_c800 had better accuracy than the original NeRF in most scenes. Note that PlenOctrees are converted to from the weights stored at some intervals during the training of NeRF-SH and then fine-tuned. Because the time required for fine tuning varies, the graph does not show a simple upward trend.}
  \label{fig:psnrs}
\end{figure*}

\subsection{Experimental Setup}
\label{sec:setup}

We detail the datasets and basic setups used in our experiments. More experimental setup details are included in Supplementary Materials.

\paragraph{Datasets}
For our experiments, we use the NeRF-Synthetic dataset~\cite{mildenhall2020nerf} and NSVF-Synthetic dataset~\cite{liu2020neural}. The NeRF-Synthetic dataset and the NSVF-Synthetic dataset each consist of 8 scenes where each scene has a central object with 100 inward facing cameras distributed randomly on the upper hemisphere. All scenes exhibit complicated geometry and realistic non-Lambertian materials, but the NSVF-Synthetic dataset has more complex geometry, lighting effects and transparent parts, as shown in~\autoref{fig: nsvf_qualitative}. The images are 800 x 800  in resolution with ground-truth camera poses.

\paragraph{Baselines}
The main baseline in our experiments is NeRF's JAX re-implementation JaxNeRF~\cite{jaxnerf2020github}.
It is slightly faster (much faster with parallel machines) to learn and inference, and easier to extend than the original implementation~\cite{mildenhall2020nerf}. NeRF has a large number of fine sampling points by default (128) that may be unfair in speed comparisons, so we run it with different numbers of fine sampling points: 16, 64, and 128. In \autoref{fig:psnrs} and \autoref{tab:psnrssim}, each NeRF result is denoted as cXfY using the number of coarse samples (X) and fine samples (Y). We also compare to two recent papers introducing NeRF accelerations: neural sparse voxel fields (NSVF) and PlenOctrees.
Each experiment is run on a single NVIDIA A100 GPU.


\paragraph{VaxNeRF Implementation}
Like NeRF, VaxNeRF is denoted as vax\_cXfY using the number of coarse samples (X) and fine samples (Y) in \autoref{fig:psnrs} and \autoref{tab:psnrssim}.
Besides a modular visual hull function, we brought down the implementation to about \numlines~lines of change on top of JaxNeRF, as shown in~\autoref{fig:commit}.
Specifically, the only changes are to read in the voxels computed by our visual hull, add the voxels as an argument to the function, and then enter only the points inside the bounding volume into the MLP, and set values of zero for all other points.
Unlike in NeRF where the number of points is fixed for all rays ($N$ in Eq.~\ref{eqn:nerfray}), in VaxNeRF it varies per ray due to rejecting points outside the bounding volume (Eq.~\ref{eqn:vaxnerfray}).
Since JAX does not efficiently handle variable-length inputs at the moment, in our current implementation we run the full NeRF model without rejection for 30 seconds, record the maximum number of points inside the volume across all rays, and compiles our code with the input sizes as  ``len\_inpc'' and ``len\_inpf'' for coarse and (optional) fine models respectively\footnote{This may not be necessary in the future, as JAX improves compilation caching.}.

\paragraph{Visual Hull Implementation}
We use 400x400x400 resolution of voxels for visual hull.
In addition, we need a binary mask separating the foreground from the background in each image.
In the scenes where the alpha channel is given, we use it to directly derive the mask, but in some scenes of the NSVF-Synthetic dataset where it is not given, we use OpenCV to separate the background (white) from the foreground. In a real environment,  one could use a green background, or apply an off-the-shelf segmentation model for separating objects of interest from the rest after capturing.

\subsection{Experimental Results}
\label{sec:exp_results}

We provide extensive evaluations on benchmark datasets against a series of strong baselines. More results are included in Supplementary Materials.

\textbf{Qualitative Comparisons}~~In~\autoref{fig: qualitative}, we show how NeRF and VaxNeRF visually compare in the first 30 minutes of training. We can observe substantial details emerge even with only 15-30 minutes of training on a single GPU. We also include the colored visualization of visual hull for reference, even though our method did not use this color information. This visual hull is optimized in the first 10 seconds, and yet it clearly captures intricate geometric details at 400x400x400 voxel resolution. 

\input{figures/nsvf_results}

Additionally, we show the final rendering quality of VaxNeRF on complex NSVF-Synthetic dataset~\cite{liu2020neural}. As our quantitative metrics below and these visualizations illustrate, VaxNeRF does not degrade quality even on those scenes with complex geometry, lighting effects and transparent parts, unlike many other prior acceleration works in~\autoref{tab:relworks}. This is because visual hull's property and VaxNeRF's design ensure that it converges to the same result as a full NeRF, as discussed in Section~\ref{sec:designchoices}.

\textbf{Quantitative Comparisons}~~\autoref{fig:psnrs} plots PSNR over training time, where all results are performances on a single NVIDIA A100 GPU. We can cleanly observe that all VaxNeRF variants (``vax\_X'') have fast convergence with high asymptotic accuracy. With sufficient point samples (vax\_c600, vax\_c800 and vax\_c64f128), our VaxNeRF performs on par with NeRF in final rendering quality while speeding up the training by~\speedup, and sometimes even outperforms NeRF on final accuracy (vax\_c600). Our results also compare favorably with results of NSVF and PlenOctrees reproduced from their official open-source implementations. NVSF, a popular model optimized for fast inference, appears slower in training since it gradually refines voxel resolutions during training. More surprisingly, compared to PlenOctrees which is substantially more complex and highly-optimized for both training and inference, VaxNeRF performs comparatively in terms of both speed and final rendering quality in all cases (except e.g. ``ship''), slightly outperforming in PSNR and little under-performing in SSIM. ~\autoref{tab:psnrssim} summarizes these results comprehensively, where VaxNeRF, along with PlenOctrees, being the most performant across all time windows of training. 


\input{tables/analysis}

Notably,~\autoref{fig:psnrs} and~\autoref{tab:psnrssim} show that the most performant VaxNeRF only needs a single coarse model with dense point sampling (e.g. vax\_600 and vax\_800), a strong advantage of VaxNeRF over NeRF discussed in Section~\ref{sec:designchoices}. Table \ref{tab:analysis} summarizes how our method dramatically reduces the number of points evaluated by NeRF, which in turn allows it to proceed with learning at a higher rays/sec. Our c\_600 and c\_800 VaxNeRF achieve training speed-ups of 5.35x and 4.11x per iteration respectively on average over a total of 16 scenes on the NeRF-Synthetic and NSVF-Synthetic datasets with little or no deterioration, or even slight improvements, in quality.


%% file: tables/results.tex
\begin{table*}
\centering
\begin{tabular}{cllllllllllll}
\multicolumn{13}{c}{Synthetic NeRF Dataset \hspace{1em} \textfirstb{best} \textsecondb{second-best} \textthirdb{third-best}} \\
\toprule
& \multicolumn{2}{c}{0.5h} & \multicolumn{2}{c}{1h} & \multicolumn{2}{c}{3h} & \multicolumn{2}{c}{6h} & \multicolumn{2}{c}{12h} & \multicolumn{2}{c}{36h} \\
Method & PSNR & SSIM & PSNR & SSIM & PSNR & SSIM & PSNR & SSIM & PSNR & SSIM & PSNR & SSIM \\
\midrule
c64f64 & 27.07 & 0.908 & 28.64 & 0.926 & 29.98 & 0.939 & 30.73 & 0.946 & 31.22 & 0.949 & 31.39 & 0.951 \\
c64f128 & 26.66 & 0.903 & 28.36 & 0.923 & 29.96 & 0.939 & 30.75 & 0.946 & 31.4 & 0.951 & \cellthird\bf{31.73} & 0.953 \\
vax\_c64f64 & 29.33 & 0.932 & 30.07 & 0.939 & 31.1 & 0.948 & 31.38 & 0.95 & 31.43 & 0.951 & 31.43 & 0.951 \\
vax\_c64f128 & 28.82 & 0.927 & 29.78 & 0.937 & 31.03 & 0.947 & \cellthird\bf{31.53} & 0.951 & \cellthird\bf{31.71} & 0.953 & \cellthird\bf{31.73} & 0.953 \\
vax\_c400 & \cellfirst\bf{30.04} & \cellfirst\bf{0.939} & \cellsecond\bf{30.73} & \cellfirst\bf{0.945} & \cellthird\bf{31.35} & 0.95 & 31.43 & 0.951 & 31.44 & 0.951 & 31.44 & 0.951 \\
vax\_c600 & \cellsecond\bf{30.0} & \cellsecond\bf{0.938} & \cellfirst\bf{30.8} & \cellfirst\bf{0.945} & \cellsecond\bf{31.72} & \cellthird\bf{0.952} & \cellsecond\bf{31.92} & \cellthird\bf{0.954} & \cellsecond\bf{31.95} & \cellthird\bf{0.955} & \cellsecond\bf{31.95} & \cellthird\bf{0.955} \\
vax\_c800 & \cellthird\bf{29.8} & \cellthird\bf{0.936} & \cellthird\bf{30.69} & \cellthird\bf{0.944} & \cellfirst\bf{31.78} & \cellsecond\bf{0.953} & \cellfirst\bf{32.1} & \cellsecond\bf{0.955} & \cellfirst\bf{32.17} & \cellsecond\bf{0.956} & \cellfirst\bf{32.18} & \cellsecond\bf{0.956} \\
nsvf & 24.96 & 0.875 & 26.04 & 0.893 & 26.91 & 0.908 & 27.52 & 0.92 & 28.04 & 0.93 & 28.17 & 0.934 \\
plenoctrees & 28.69 & 0.933 & 29.81 & \cellthird\bf{0.944} & 31.28 & \cellfirst\bf{0.955} & 31.47 & \cellfirst\bf{0.957} & 31.55 & \cellfirst\bf{0.957} & 31.53 & \cellfirst\bf{0.958} \\

\bottomrule
\end{tabular}
\caption{Quantitative results on NeRF-Synthetic dataset. Each value is the average of the eight scenes. VaxNeRF has a higher score than the original NeRF at all times, and the final performance is also better than the original.}
\label{tab:psnrssim}
\end{table*}

%% file: figures/hist.tex
\begin{figure*}[t]
  \def \factor {0.16}
  \def \vertspace {0cm}
  \def \horizontalspace {0cm}
  \centering
  \begin{tabular}{c@{\hspace{\horizontalspace}}|c@{\hspace{\horizontalspace}}c@{\hspace{\horizontalspace}}c@{\hspace{\horizontalspace}}c@{\hspace{\horizontalspace}}c}
    &
    \includegraphics[trim=30px 70px 50px 50px, clip, width=\factor\linewidth]{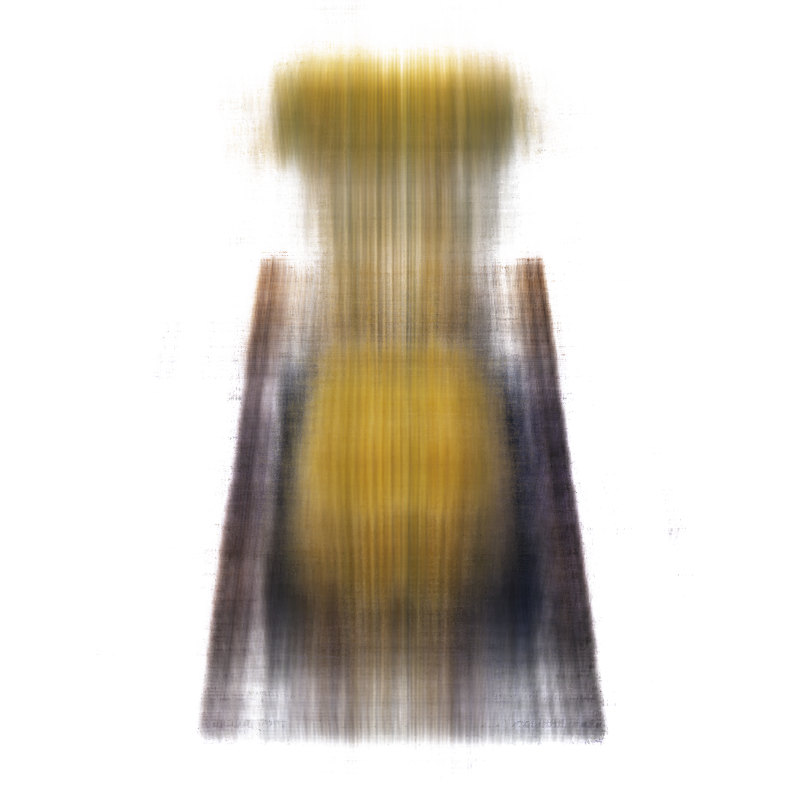} &
    \includegraphics[trim=30px 70px 50px 50px, clip, width=\factor\linewidth]{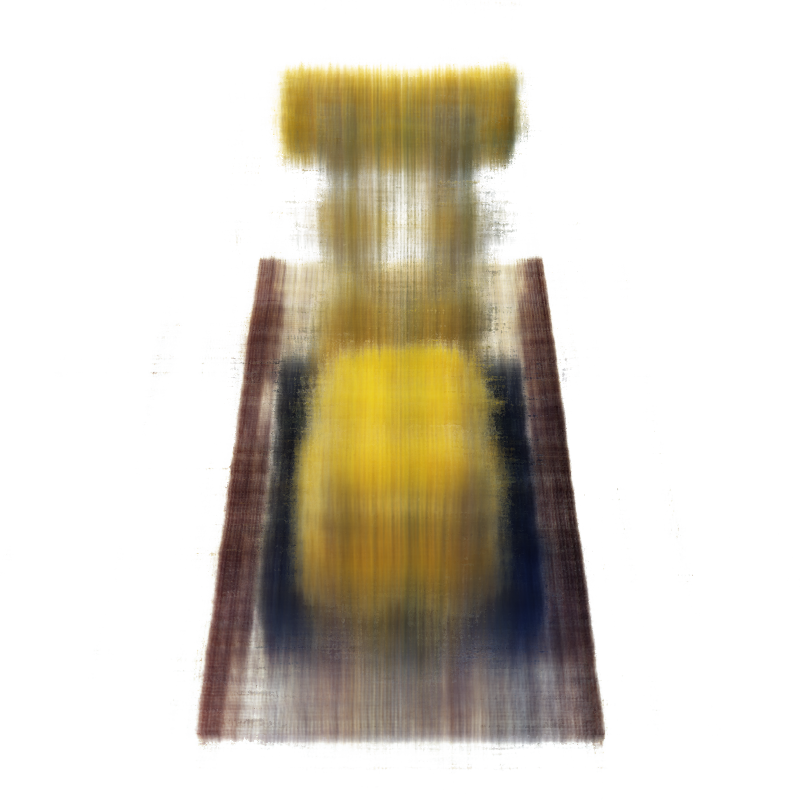} &
    \includegraphics[trim=30px 70px 50px 50px, clip, width=\factor\linewidth]{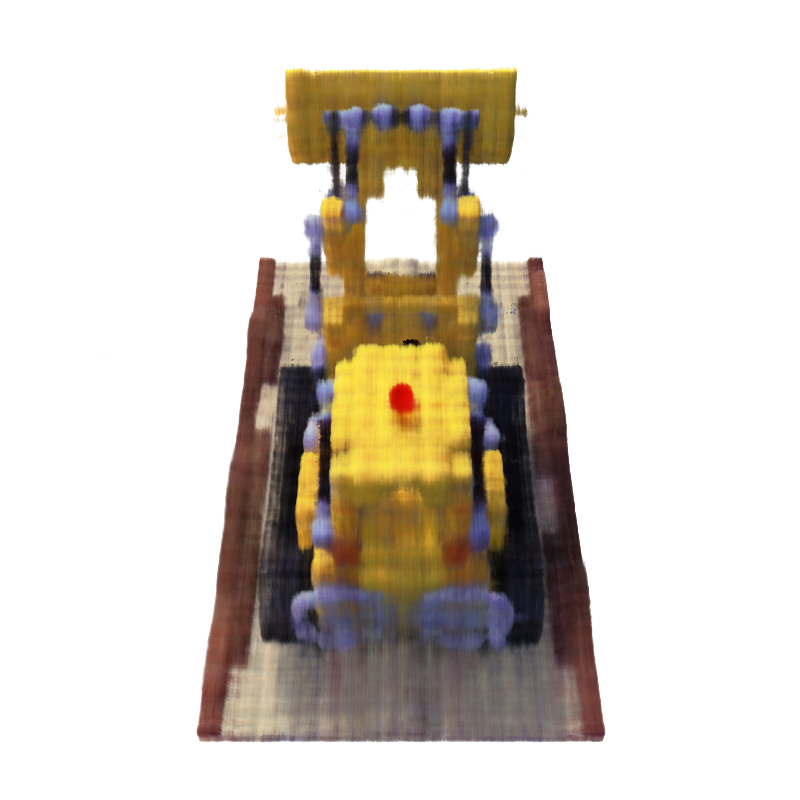} &
    \includegraphics[trim=30px 70px 50px 50px, clip, width=\factor\linewidth]{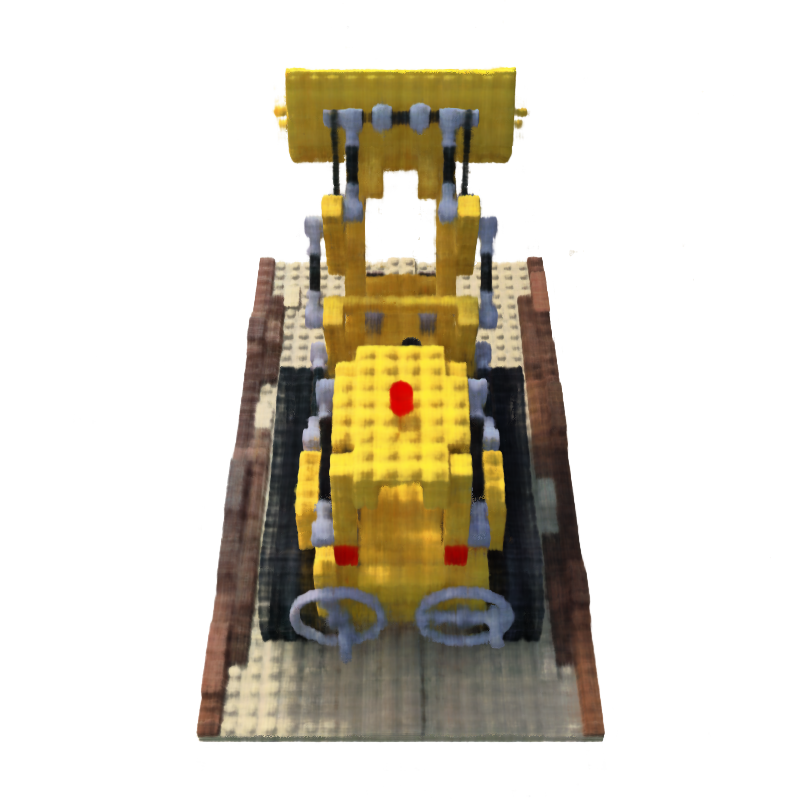} &
    \includegraphics[trim=30px 70px 50px 50px, clip, width=\factor\linewidth]{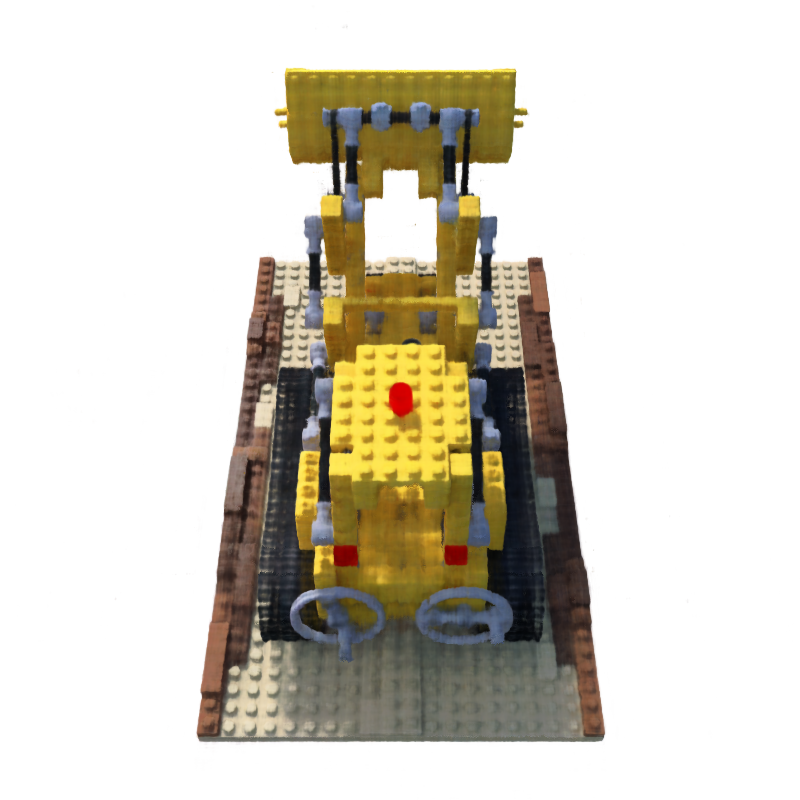}
    \vspace{\vertspace}\\

    \includegraphics[trim=30px 70px 50px 50px, clip, width=\factor\linewidth]{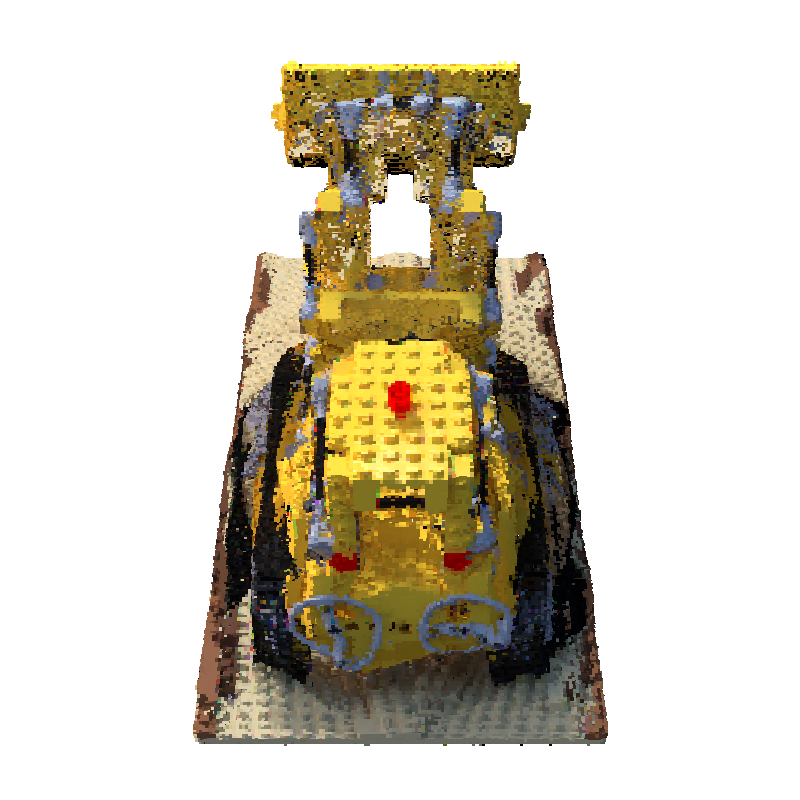} &
    \includegraphics[trim=30px 70px 50px 50px, clip, width=\factor\linewidth]{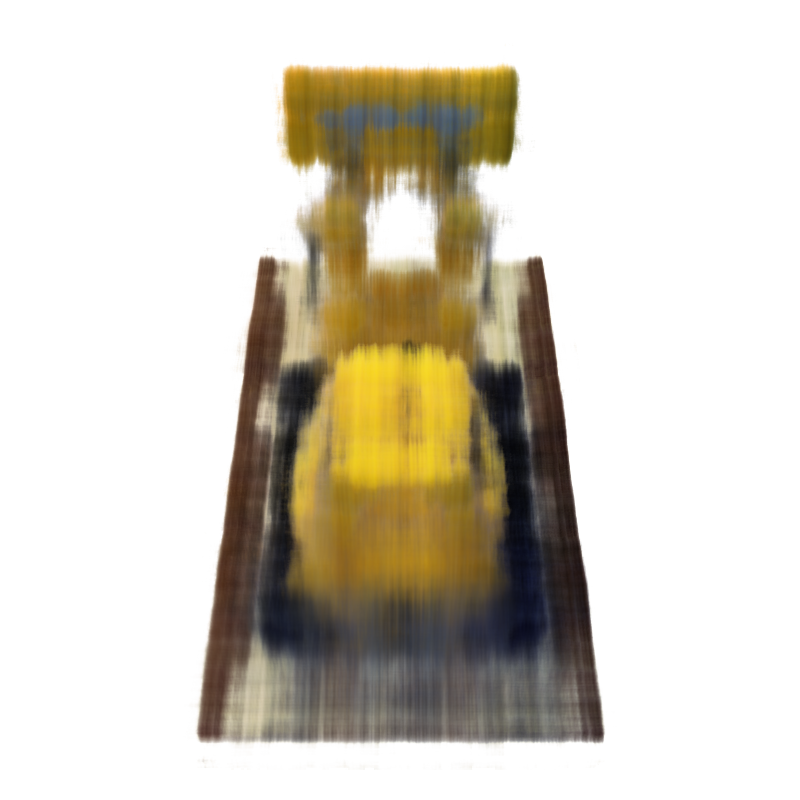} &
    \includegraphics[trim=30px 70px 50px 50px, clip, width=\factor\linewidth]{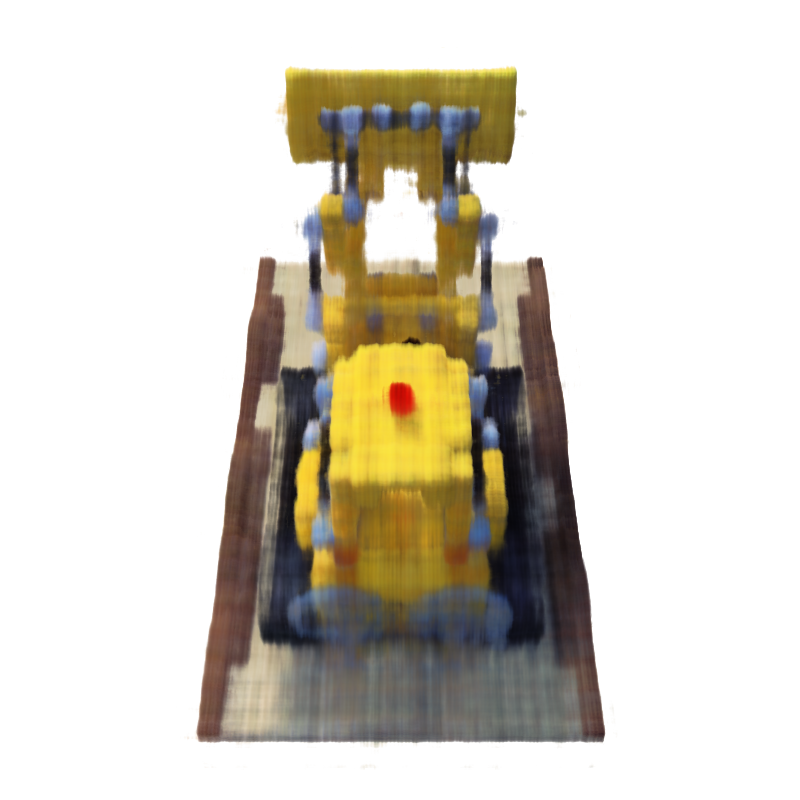} &
    \includegraphics[trim=30px 70px 50px 50px, clip, width=\factor\linewidth]{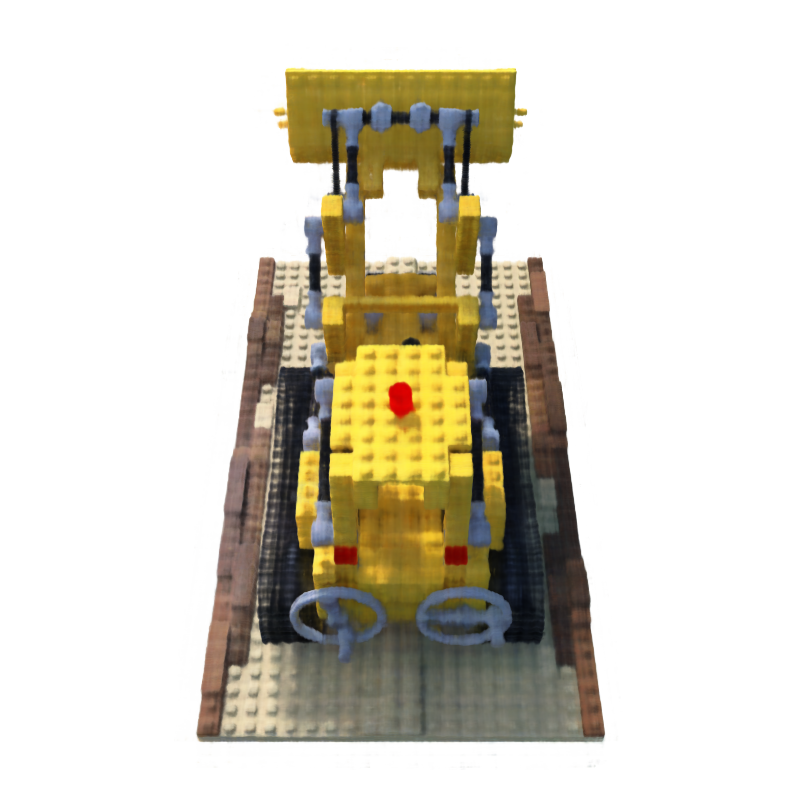} &
    \includegraphics[trim=30px 70px 50px 50px, clip, width=\factor\linewidth]{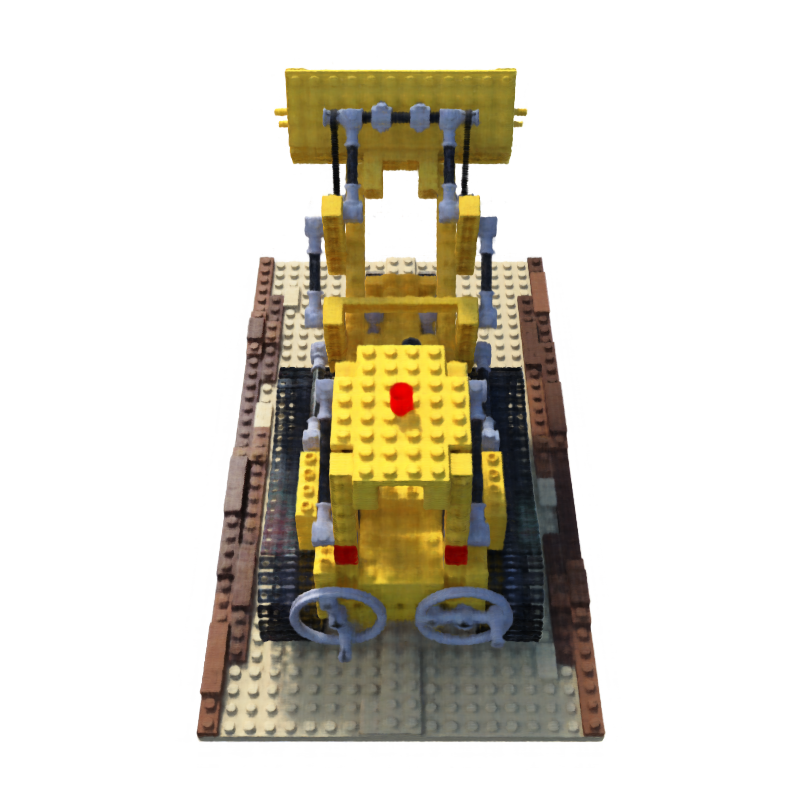} &
    \includegraphics[trim=30px 70px 50px 50px, clip, width=\factor\linewidth]{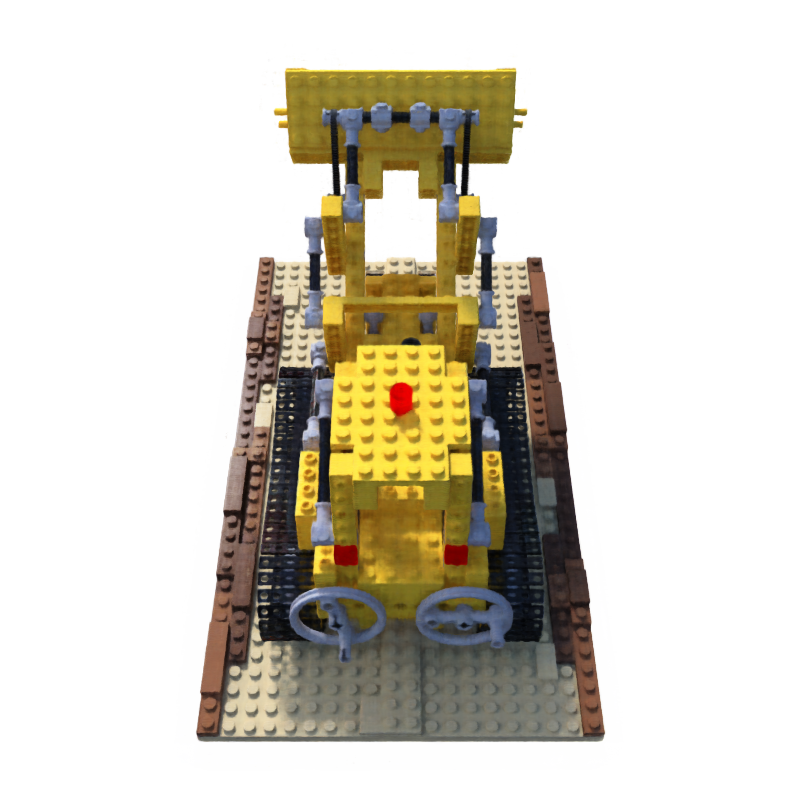}
    \vspace{\vertspace}\\

    Visual Hull & 0.5min & 1min & 5min & 15min & 30min \\
  \end{tabular}
  \caption{Top row is (Jax)NeRF (c64f128) results, bottom row is VaxNeRF (vax\_c600) results. The elapsed time of VaxNeRF includes the time taken by Visual Hull. (Ten seconds, though.) Thanks to the visual hull, VaxNeRF is able to acquire the shape very fast. For example, VaxNeRF can already start learning the smallest holes in the caterpillar in 5 minutes, while the original NeRF can only do so after about 30 minutes. Please see Supplementary Materials for more visualizations, including zoomed comparisons along longer time scales. Note that, the colors (textures) shown in the ``Visual Hull'' are not used for VaxNeRF training.}
  \label{fig: qualitative}

\end{figure*}

%% file: figures/nsvf_results.tex
\begin{figure}[t]
  \def \factor {0.32}
  \def \vertspace {0cm}
  \def \horizontalspace {0cm}
  \centering
  \begin{tabular}{c@{\hspace{\horizontalspace}}c@{\hspace{\horizontalspace}}c@{\hspace{\horizontalspace}}c@{\hspace{\horizontalspace}}c}
    \includegraphics[trim=90px 140px 180px 130px, clip, width=\factor\linewidth]{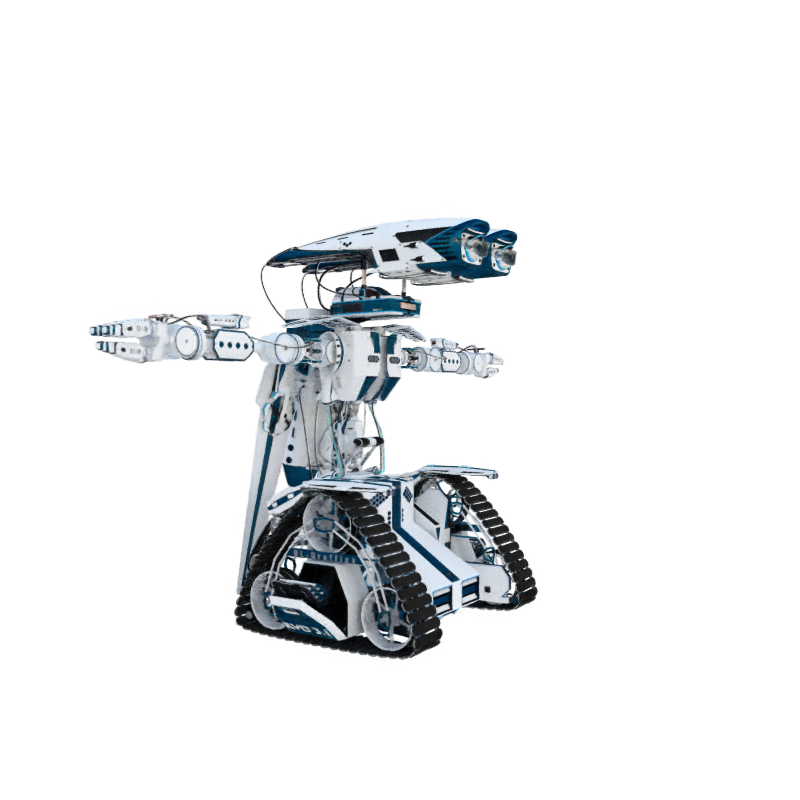} &
    \includegraphics[trim=210px 200px 120px 130px, clip, width=\factor\linewidth]{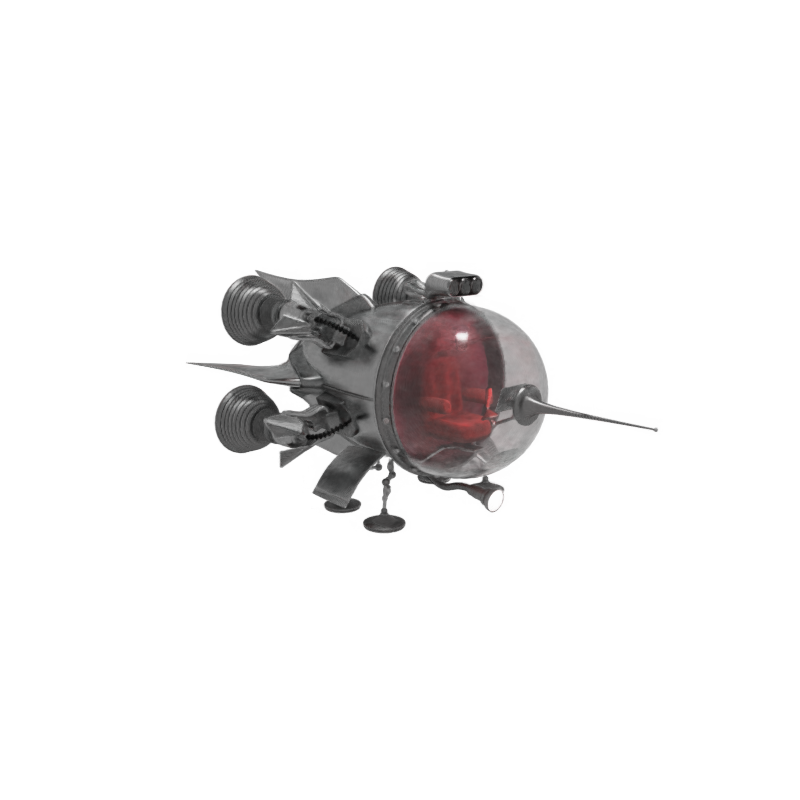} &
    \includegraphics[trim=220px 180px 0px 0px, clip, width=\factor\linewidth]{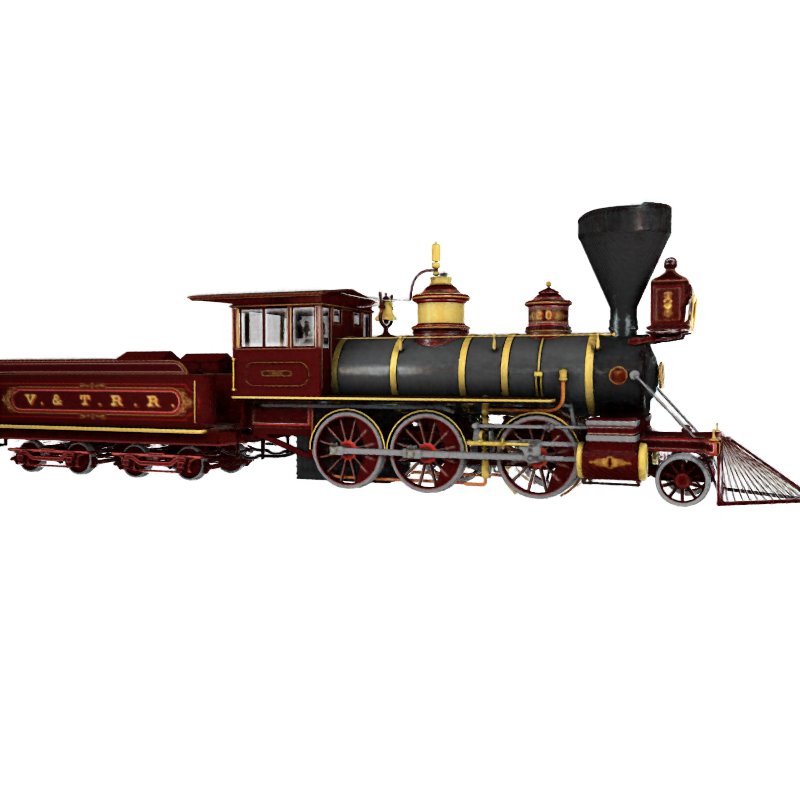}
    \vspace{\vertspace}\\

    \includegraphics[trim=75px 50px 75px 100px, clip, width=\factor\linewidth]{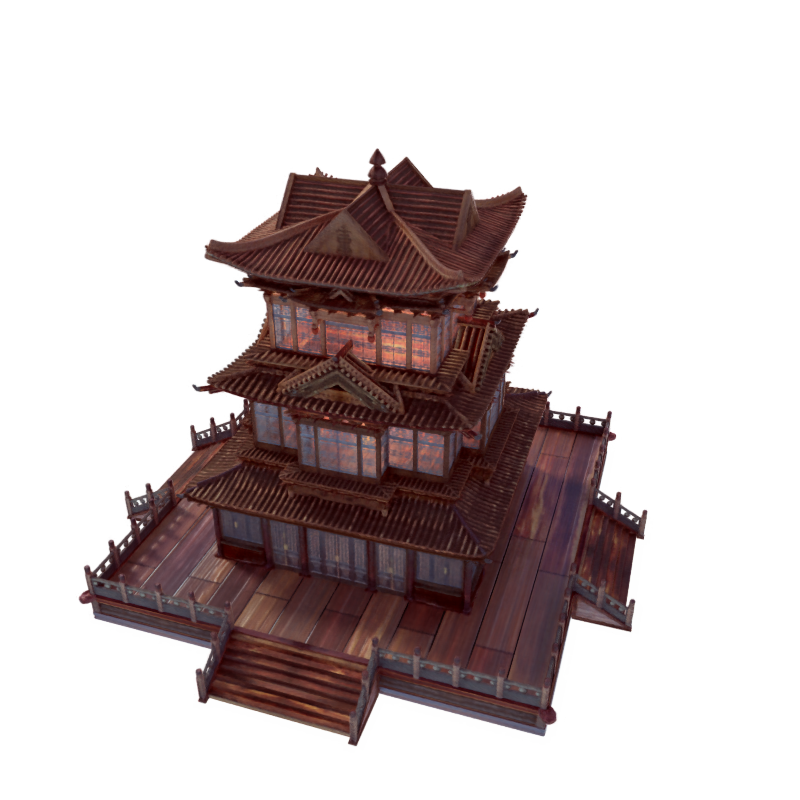} &
    \includegraphics[trim=200px 180px 40px 80px, clip, width=\factor\linewidth]{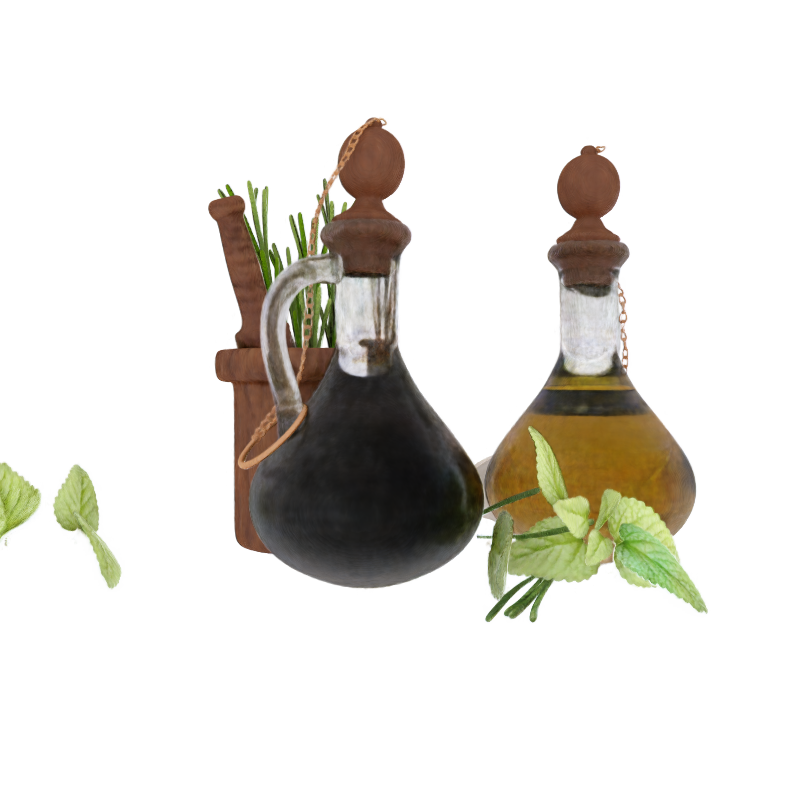} &
    \includegraphics[trim=50px 0px 50px 100px, clip, width=\factor\linewidth]{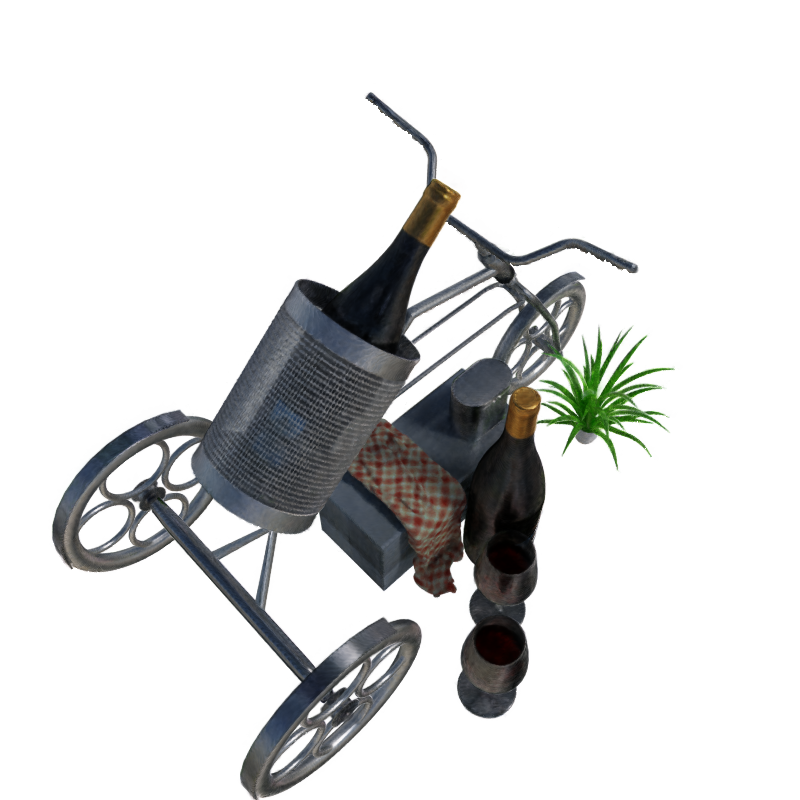}
    \vspace{\vertspace}\\

  \end{tabular}
  \caption{Qualitative results of VaxNeRF (vax\_c600) on NSVF-Synthetic datasets, after 1 million training iterations, averaging 5.6 hours (this takes 33 hours on NeRF). VaxNeRF also performs well on objects with transparency and finer shapes.}
  \label{fig: nsvf_qualitative}

\end{figure}

%% file: tables/analysis.tex
\begin{table}
  \centering
  \begin{tabular}{@{}ccc@{}}
    \toprule
    Method & number of samples & rays / sec \\
    \midrule
    c64f128 & 802,816 & 34,424 \\
    c64f64 & 524,288 & 45,153 \\
    c64f16 & 327,680 & 58,610 \\
    vax\_c64f128 & 363,625 & 89,151 \\
    vax\_c64f64 & 211,625 & 141,086 \\
    vax\_c64f16 & 97,875 & 240,489 \\
    vax\_c800 & 184,500 & 148,812 \\
    vax\_c600 & 138,500 & 192,489 \\
    vax\_c400 & 92,125 & 275,824 \\
    \bottomrule
  \end{tabular}
  \caption{Training Speed Comparison on NeRF Synthesic Dataset. Measured on a single NVIDIA A100 GPU. Although vax\_400, vax\_600, and vax\_800 sample the points on the ray quite finely in the voxel, the overall number of points to be sampled is dramatically reduced, and the training proceeds at a high ray/sec.}
  \label{tab:analysis}
\end{table}


%% file: sections/6_conclusion.tex
\section{Discussion}
\label{sec:discussion}
While our empirical experiments conclusively demonstrate the speed-ups of VaxNeRF over NeRF, in this section we discuss possible explanations for such substantial gains, specifically:
\begin{itemize}
    \item \textbf{Effective Resolution for Point Selection}: Both NeRF and VaxNeRF use heuristics to better allocate points for integral estimation along each ray (Eq.~\ref{eqn:nerfray}). Given that NeRF uses 64 samples for coarse model evaluations and VaxNeRF fits 400x400x400 visual hull,
    these correspond to effective resolutions of $1/64$ and $1/400$ respectively for selecting important regions along the ray.  
    Despite NeRF using more direct estimate of contribution ($w_i$ instead of 0/1 masking) and visual hull's imperfect boundary volume carving, this large effective resolution gap could explain VaxNeRF's superior performance.
    \item \textbf{Zero-Region Predictions}: In NeRF, the 3D space regions where no object is present are constantly evaluated and optimized throughout the training for both coarse and fine models, while in VaxNeRF they are directly set to 0 without ever running MLP evaluations. This substantially could reduce the predictive burden on MLPs in VaxNeRF, leading to faster training.
\end{itemize}
We speculate that the overall speed-up arises from a combination of these design effects, and hope more future works probe into the fundamental building blocks of NeRF.

\section{Conclusion}
\label{sec:conclusion}

We present VaxNeRF, combining the classic visual hull technique with NeRF. Our implementation only takes about \numlines~lines of main code changes from JaxNeRF, and yet provides reliable \speedup~speed-ups compared to this strong baseline with no loss in rendering quality. We hope our results could encourage more works
asking fundamental questions about algorithmic designs of the popular end-to-end methods, and revisiting classic approaches (that are supposedly replaced by these methods) within these methods.

\paragraph{Limitations and Future Work} The limitations of the current work are: (1) visual hull requires binary foreground-background masks, and (2) all results are on synthetic datasets. Evaluating VaxNeRF on real-world datasets are the next challenge, where we plan to curate and open-source such a dataset with ground-truth masks from green screen or an off-the-shelf segmentation model.

%% file: sections/9_appendix.tex
\section{Additional Experimental Setups}
\label{sec:appendix_exp_setups}

This table shows hyper-parameters we used for experiments. For NSVF and PlenOctrees, we ran the official implementation as is.

\input{tables/hyperparams}

\begin{algorithm*}[t]
    \caption{NeRF and VaxNeRF (w/o hierarchical sampling). \textcolor{red}{Added codes} to NeRF and \textcolor{blue}{removed codes} from NeRF}
    \label{alg:nerf}
    \begin{algorithmic}[1]
        \REQUIRE dataset of rays and corresponding color $\mathcal{D}$
            \WHILE {not done}
                \STATE Sample batch of rays and color $\{\{o_i,d_i\}, \Col_i\} \sim \mathcal{D}$
                \STATE Coarse sampling points along the ray $\mathbf{x}_i^{(c)} \Leftarrow \text{coarse\_sample}(o_i, d_i)$
                \STATE \textcolor{blue}{Evaluate color and density at each point $\mathbf{c}_i^{(c)}, \mathbf{\sigma}_i^{(c)} = \text{MLP}_{\mathbf{\theta}_c}(\mathbf{x}_i^{(c)}, d_i)$}
                \STATE \textcolor{red}{Evaluate color and density at each point $\mathbf{c}_i^{(c)}, \mathbf{\sigma}_i^{(c)} \Leftarrow \left\{\begin{matrix}
                \text{MLP}_{\mathbf{\theta}_c}(\mathbf{x}_i^{(c)}, d_i)\quad&(\mathbf{x}_i\in \vh)\\
                \mathbf{0}\quad&(\mathbf{x}_i\notin \vh)
                \end{matrix}\right. $}
                \STATE \textcolor{blue}{Compute pixel value using ray marching $\estimate{\Col}_i^{(c)} \Leftarrow \text{ray\_marching}(\mathbf{x}_i^{(c)}, \mathbf{c}_i^{(c)}, \mathbf{\sigma}_i^{(c)})$}
                \STATE \textcolor{blue}{fine sampling points along the ray $\mathbf{x}_i^{(f)} \Leftarrow \text{fine\_sample}(o_i, d_i)$}
                \STATE \textcolor{blue}{Evaluate color and density at each point $\mathbf{c}_i^{(f)}, \mathbf{\sigma}_i^{(f)} \Leftarrow \text{MLP}_{\mathbf{\theta}_f}(\mathbf{x}_i^{(f)}, d_i)$}
                \STATE \textcolor{blue}{Compute pixel value using ray marching $\estimate{\Col}_i^{(f)} \Leftarrow \text{ray\_marching}(\mathbf{x}_i^{(f)}, \mathbf{c}_i^{(f)}, \mathbf{\sigma}_i^{(f)})$}
                
                \STATE Compute loss for each pixel $\mathcal{L}_c = \sum \big\lVert{\truecol_i - \estimate{\Col}_i^{(c)}}\big\rVert^2_2$, \textcolor{blue}{$\mathcal{L}_f = \sum \big\lVert{\truecol_i - \estimate{\Col}_i^{(f)}}\big\rVert^2_2$}
                \STATE Update parameters $\mathbf{\theta}_c$, \textcolor{blue}{$\mathbf{\theta}_f$} using gradient descent
            \ENDWHILE
    \end{algorithmic}
\end{algorithm*}

\section{Additional Experimental Results}
\label{sec:appendix_exp_results}

\begin{figure*}
\centering
  \includegraphics[width=1.\linewidth]{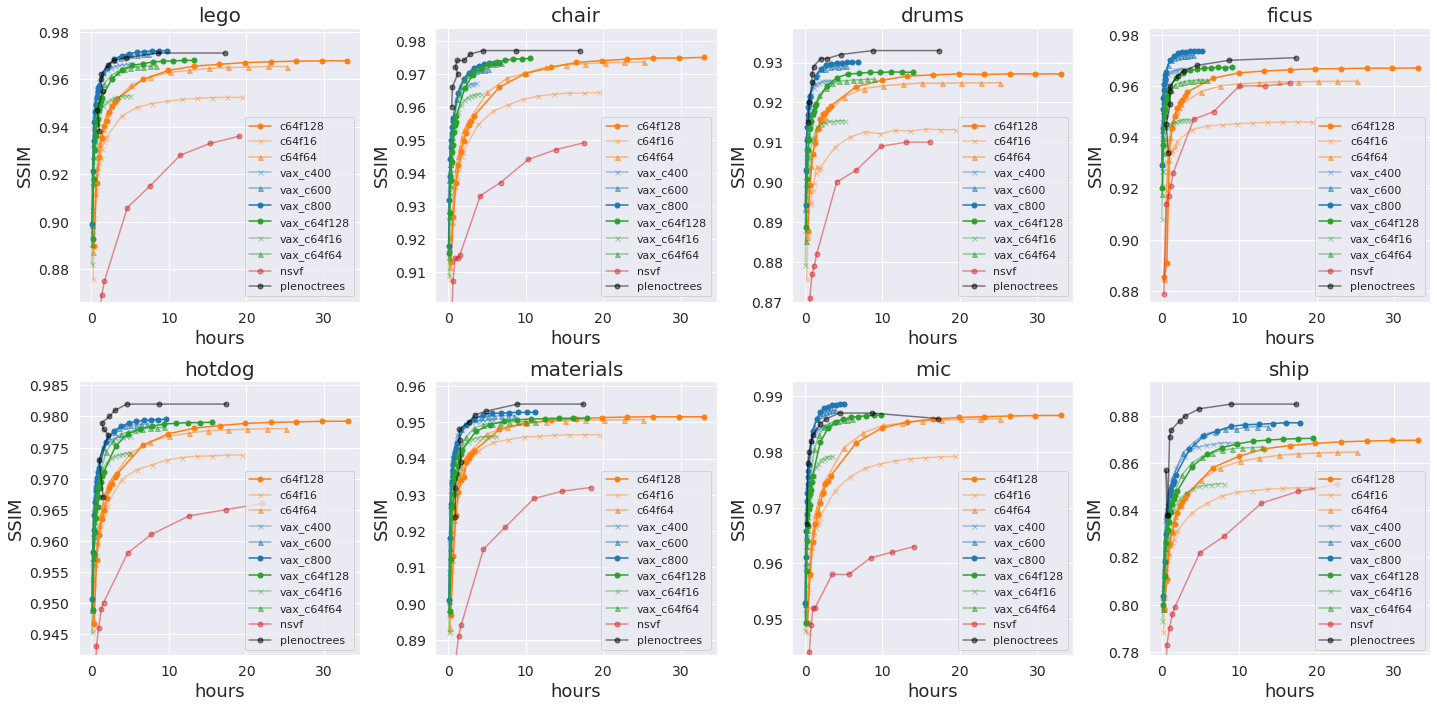}
  \caption{SSIM comparison on NeRF-Synthetic dataset}
  \label{fig:ssims}
\end{figure*}

\input{tables/results_nsvfdata}

\begin{figure*}
  \centering
  \includegraphics[width=1.\linewidth]{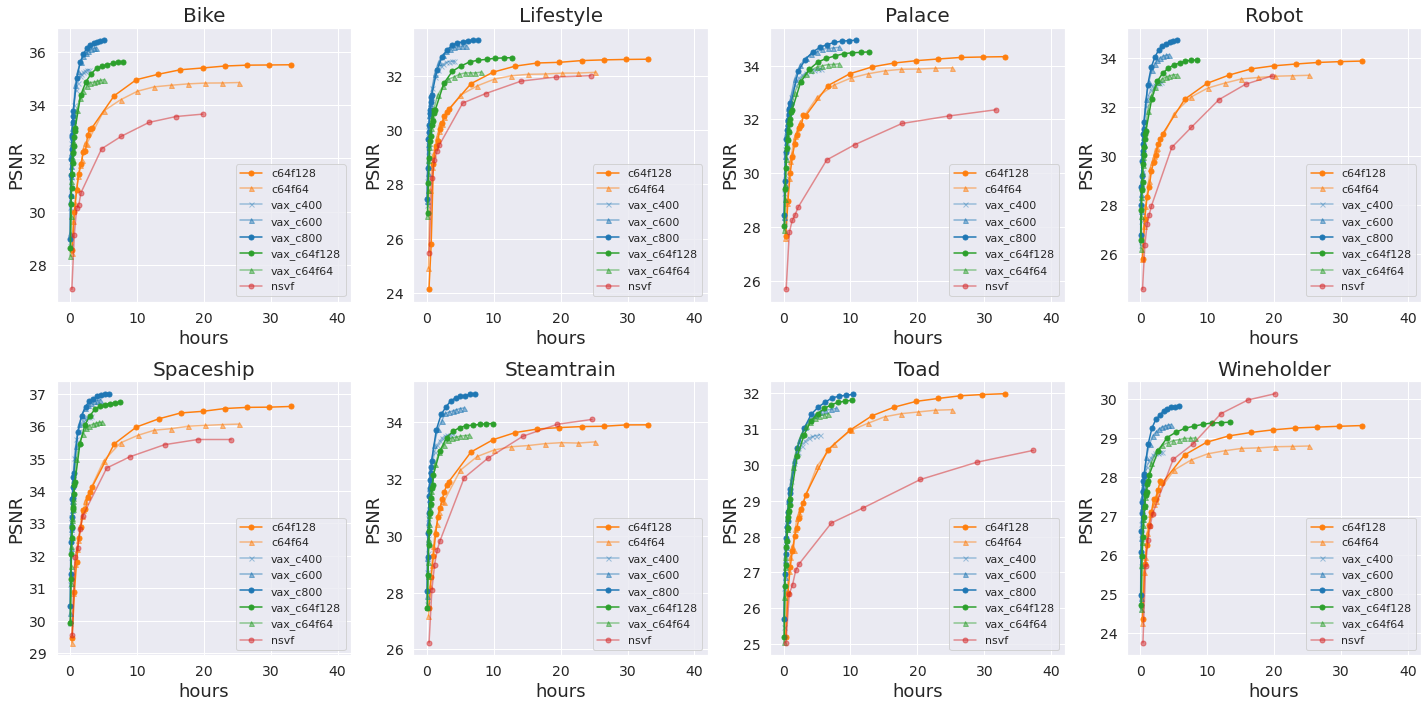}
  \caption{PSNR comparison on NSVF-Synthetic dataset}
  \label{fig:psnrsnsvf}
\end{figure*}

\begin{figure*}
  \centering
  \includegraphics[width=1.\linewidth]{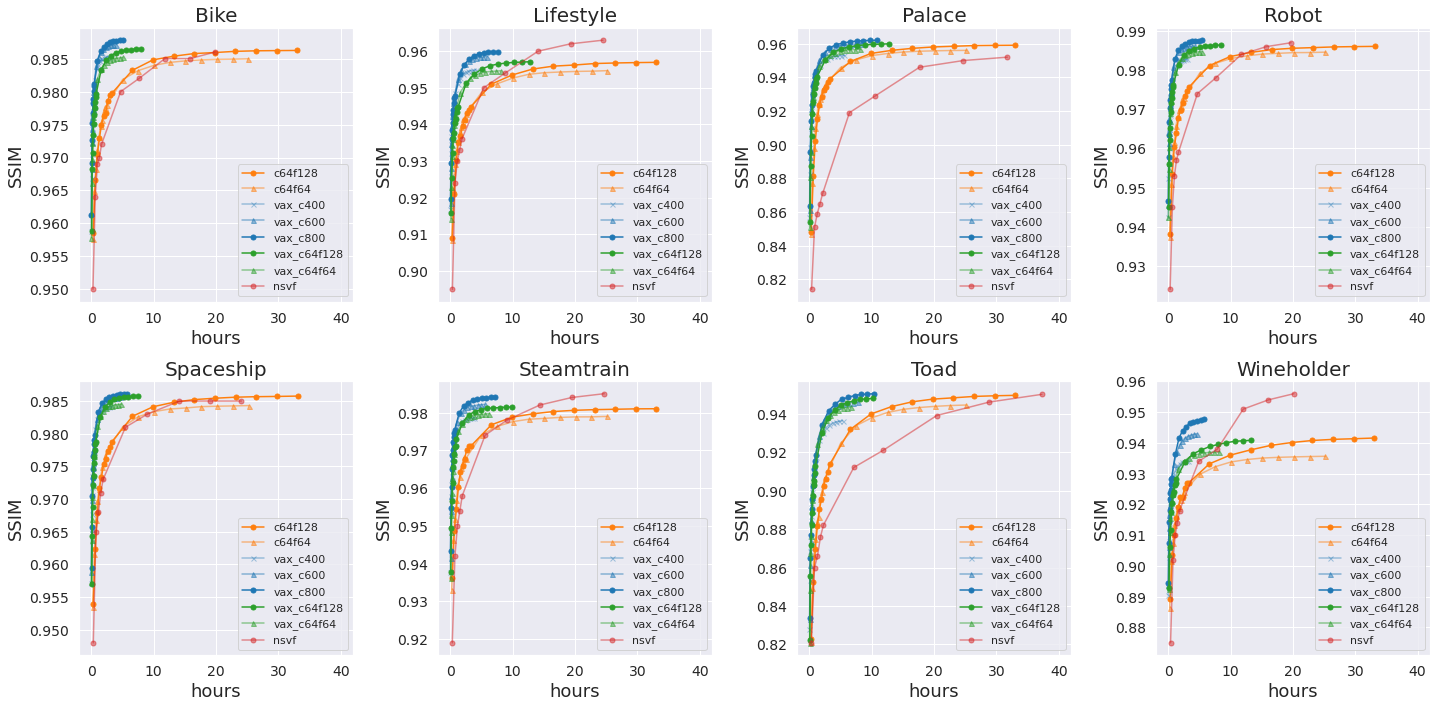}
  \caption{SSIM comparison on NSVF-Synthetic dataset}
  \label{fig:ssimsnsvf}
\end{figure*}


%% file: tables/hyperparams.tex
\begin{table}[h]
  \centering
  \begin{tabular}{@{}cccc@{}}
    \toprule
    Method & batch\_size & max\_iteration & emb\_dim \\
    \midrule
    (Jax)NeRF & 4,096 & 1,000,000 & 10 \\
    VaxNeRF & 4,096 & 1,000,000 & 10 \\
    NSVF & 4,096 & 150,000 & 10 \\
    PlenOctrees & 1,024 & 2,000,000 & 10 \\
    \bottomrule
  \end{tabular}
  \caption{Hyper-parameters of each method.}
  \label{tab:hyper}
\end{table}

%% file: tables/results_nsvfdata.tex
\begin{table*}
\centering
\begin{tabular}{cllllllllllll}
\multicolumn{13}{c}{Synthetic NSVF Dataset \hspace{1em} \textfirstb{best} \textsecondb{second-best} \textthirdb{third-best}} \\
\toprule
& \multicolumn{2}{c}{0.5h} & \multicolumn{2}{c}{1h} & \multicolumn{2}{c}{3h} & \multicolumn{2}{c}{6h} & \multicolumn{2}{c}{12h} & \multicolumn{2}{c}{36h} \\
Method & PSNR & SSIM & PSNR & SSIM & PSNR & SSIM & PSNR & SSIM & PSNR & SSIM & PSNR & SSIM \\
\midrule
c64f64 & 28.05 & 0.922 & 29.32 & 0.936 & 31.03 & 0.952 & 32.06 & 0.959 & 32.74 & 0.964 & 32.98 & 0.965 \\
c64f128 & 27.3 & 0.916 & 29.06 & 0.933 & 31.16 & 0.953 & 32.17 & 0.96 & 33.03 & 0.965 & 33.52 & 0.968 \\
vax\_c64f64 & 30.61 & 0.947 & 31.58 & 0.955 & 32.8 & \cellthird\bf{0.964} & 33.05 & 0.966 & 33.07 & 0.966 & 33.07 & 0.966 \\
vax\_c64f128 & 30.21 & 0.943 & 31.36 & 0.953 & 32.89 & \cellthird\bf{0.964} & \cellthird\bf{33.42} & \cellthird\bf{0.967} & \cellthird\bf{33.57} & \cellthird\bf{0.968} & \cellthird\bf{33.58} & 0.968 \\
vax\_c400 & \cellsecond\bf{31.22} & \cellsecond\bf{0.951} & \cellthird\bf{32.16} & \cellthird\bf{0.957} & \cellthird\bf{32.95} & 0.963 & 33.02 & 0.964 & 33.02 & 0.964 & 33.02 & 0.964 \\
vax\_c600 & \cellfirst\bf{31.33} & \cellfirst\bf{0.952} & \cellfirst\bf{32.33} & \cellfirst\bf{0.959} & \cellsecond\bf{33.56} & \cellsecond\bf{0.967} & \cellsecond\bf{33.77} & \cellsecond\bf{0.968} & \cellsecond\bf{33.79} & \cellsecond\bf{0.969} & \cellsecond\bf{33.79} & \cellthird\bf{0.969} \\
vax\_c800 & \cellthird\bf{31.19} & \cellthird\bf{0.95} & \cellsecond\bf{32.28} & \cellfirst\bf{0.959} & \cellfirst\bf{33.71} & \cellfirst\bf{0.968} & \cellfirst\bf{34.08} & \cellfirst\bf{0.97} & \cellfirst\bf{34.15} & \cellfirst\bf{0.971} & \cellfirst\bf{34.15} & \cellfirst\bf{0.971} \\
nsvf & 26.68 & 0.902 & 28.41 & 0.923 & 29.82 & 0.939 & 31.03 & 0.952 & 31.92 & 0.962 & 32.68 & \cellsecond\bf{0.97} \\
\bottomrule
\end{tabular}
\caption{Quantitative results on NSVF-Synthetic dataset. Each value is the average of the eight scenes.}
\label{tab:psnrssim_nsvfdata}
\end{table*}